\documentclass[letterpaper,fleqn]{cas-dc}

\usepackage[numbers,sort&compress]{natbib}
\usepackage{amsmath}
\usepackage{amssymb}

\usepackage{booktabs}
\usepackage{tabularx}
\usepackage{array}
\usepackage{multirow}
\usepackage{makecell}
\usepackage{hhline}
\usepackage{adjustbox}
\usepackage{colortbl}
\newcolumntype{Y}{>{\centering\arraybackslash}X}

\usepackage{graphicx}
\usepackage{rotating}

\usepackage{subcaption}

\usepackage[ruled,vlined,linesnumbered]{algorithm2e}

\usepackage{placeins}
\usepackage{multicol}
\usepackage[switch]{lineno}
\usepackage{fancyhdr}

\usepackage{times}
\usepackage{latexsym}
\usepackage{inconsolata}
\usepackage{microtype}
\usepackage{xspace}

\usepackage[T1]{fontenc}
\usepackage[utf8]{inputenc}

\usepackage{soul}
\usepackage[normalem]{ulem}
\usepackage{xcolor}

\usepackage{enumerate}
\usepackage{enumitem} 
\usepackage{etoolbox}

\usepackage{hyperref}

\renewcommand{\ttdefault}{ptm}

\def\tsc#1{\csdef{#1}{\textsc{\lowercase{#1}}\xspace}}
\tsc{WGM}
\tsc{QE}
\tsc{EP}
\tsc{PMS}
\tsc{BEC}
\tsc{DE}

\makeatletter
\patchcmd{\thebibliography}
  {\clubpenalty4000\widowpenalty4000}
  {\clubpenalty4000\widowpenalty4000\color{RoyalBlue}}
  {}{}
\makeatother

\begin{document}

\let\WriteBookmarks\relax

\shorttitle{Medical Image Analysis}
\shortauthors{M. N. Hasan et al.}

\title[mode = title]{DL\textsuperscript{3}M: A Vision-to-Language Framework for Expert-Aligned Clinical Reasoning through Deep Learning and Large Language Models}

\author[1]{Md. Najib Hasan}
\ead{mxhasan39@shockers.wichita.edu}
\affiliation[1]{organization={School of Computing}, addressline={Wichita State University}, city={Wichita}, postcode={67260}, state={Kansas}, country={USA}}

\author[2]{Imran Ahmad}
\ead{ixahmad1@shockers.wichita.edu}
\affiliation[2]{organization={W. Frank Barton School of Business}, addressline={Wichita State University}, city={Wichita}, postcode={67260}, state={Kansas}, country={USA}}

\author[3]{Sourav Basak Shuvo}
\ead{shuvo1815010@stud.kuet.ac.bd, sourav.basak.kuet.bme@gmail.com}
\affiliation[3]{organization={Department of Biomedical Engineering}, addressline={Khulna University of Engineering \& Technology}, city={Khulna}, postcode={9203}, country={Bangladesh}}

\author[4]{Md Mahadi Hasan Ankon}
\ead{0424058012@grad.cse.buet.ac.bd}
\affiliation[4]{organization={Department of Computer Science and Engineering}, addressline={Bangladesh University of Engineering and Technology}, city={Dhaka}, postcode={1000}, country={Bangladesh}}


\author[5]{Nazmul Siddique}
\ead{nh.siddique@ulster.ac.uk}
\affiliation[5]{organization={School of Computing, Engineering and Intelligent Systems}, addressline={Ulster University}, city={Londonderry}, postcode={BT48 7JL}, country={UK}}

\author[6]{Hui Wang}
\ead{h.wang@qub.ac.uk}
\cormark[1]
\affiliation[6]{organization={School of Electronics, Electrical Engineering and Computer Science}, addressline={Queen’s University Belfast}, city={Belfast}, postcode={BT9 5BN}, country={UK}}

\cortext[cor1]{Corresponding author}

\begin{abstract}
    Medical image classifiers detect gastrointestinal diseases well but do not explain their decisions. Large language models can generate clinical text, yet they struggle with visual reasoning and often produce unstable or incorrect explanations. This leaves a gap between what a model sees and the type of reasoning a clinician expects. We introduce a framework that links image classification with structured clinical reasoning. A new hybrid model, MobileCoAtNet, is designed for endoscopic images and achieves high accuracy in eight classes related to the stomach. Then, its outputs are used to drive reasoning by several LLMs. To judge this reasoning, we build two expert-verified benchmarks covering causes, symptoms, treatment, lifestyle, and follow-up care. Thirty-two LLMs are evaluated against these gold standards. Strong classification improves the quality of their explanations, but none of the evaluated models achieves consistently high clinical reasoning alignment (CRA) across prompt settings. Even the best LLMs change their reasoning when the prompts vary. Our study shows that combining DL with LLMs can produce useful clinical narratives, but current LLM remain unreliable for high-stakes medical decisions. The framework provides a clearer view of their limits and a path to build safer reasoning systems. The complete source code and datasets used in this study are available at \href{https://github.com/souravbasakshuvo/DL3M}{https://github.com/souravbasakshuvo/DL3M}. 
\end{abstract}


\maketitle

\section{Introduction}

    The Gastrointestinal (GI) system is vital for digesting food and maintaining health through organs like the stomach and intestines. However, this system is susceptible to disruptive disorders like polyps, colon cancer, and ulcerative colitis, which can affect organs such as the stomach, intestines, liver, and pancreas \cite{csener2025automatic}.  If left unmanaged, some of these abnormalities can ultimately develop into cancer \cite{ wang2025impact}.

    GI cancers represent a major global health challenge, accounting for approximately 4.8 million new cases and 3.2 million deaths in 2022, which together constitute nearly one third of all cancer related mortalities worldwide \cite{5}. A considerable proportion of this burden lies in Asia, with East Asia alone reporting 1,469,225 new gastrointestinal cancer cases and 837,360 deaths in the same year, corresponding to 43.1 percent and 41.7 percent of the global totals, respectively \cite{3}.
    Doctors use several tools to examine and treat problems in the GI tract. These include colonoscopy, upper GI endoscopy, capsule endoscopy, radiographic imaging, and lab tests \cite{ kulinna2024role}. However, diagnosis results can vary depending on the doctor's experience, which makes the process unreliable at times. Reviewing thousands of endoscopic images manually is also a major challenge, increasing the chance of missed diagnoses and delays in treatment \cite{xie2024deep,fernandes2024evaluation}, leading to conditions that can quickly progress into serious stages.
    
    Endoscopic image analysis faces issues like heavy visual clutter, inconsistent lighting, overlapping tissue patterns, and unwanted artifacts \cite{wang2026improving, hayat2025utility, lin2024evaluation}. Many diseases from different categories appear very similar, while the same disease can show different patterns in different patients. These factors reduce the accuracy of diagnosis \cite{zeng2024successful}. The quality and detail of the captured endoscopic images also play a big role in whether a condition is correctly identified \cite{tomazic20213d, mccafferty2018update}.

    To improve diagnosis, several classification methods have been developed. These methods can classify images with better precision, helping doctors in detecting and diagnosing conditions more efficiently.
    Moreover, in underdeveloped and developing countries ,the doctor-to-patient ratio is quite iimbalanced \cite{karan2021size, nawaz2025nexus}. As a result, doctors can not pay enough attention to low-risk medical conditions like stomach infections (naming of the disease) \cite{SU2025}. This raise a serious concern to develop an automated system that could reduce the workload of doctors \cite{GUMILAR20251140, GHORBIAN202585}, average patient waiting time, and increase reliability \cite{he2020deep, yilmaz2024advancing}. The automated system should be able to collect, process, classify, and give reasoning for normal and abnormal findings to serve as a reliable medical tool \cite{ELOGRI2026108744}. This would lead to faster, more accurate treatment decisions and help lower the number of deaths from GI diseases worldwide \cite{john2024advancements}.
    
    Over the past few years,  a significant number of studies have been conducted on the classification of stomach infections using deep learning \cite{NASEEM20252481, SIDDIQUI2025100304, 10534480}. Several studies have explored the usability of large language models (LLMs) in medical domain for clinical question answering, report generation, and decision support. In some studies, LLMs have been evaluated for medical text understanding, diagnosis, and prediction from textual data. However, none of these studies combined image-based disease classification with reasoning tasks. 

    In our experiment, we explore the effectiveness of LLMs for deep learning model assisted reasoning generation. This approach is necessary as LLMs are text-based models and cannot directly interpret the visual data, such as endoscopic images. If we send image in a different format (like a NumPy array or a matrix) to a text-based LLM using prompt, there will be a large information gap and the model will fail to extract important features from that image. Our solution incorporates DL models to first detect the disease from images, and then uses LLMs to provide appropriate reasoning and initial clinical advice on the diagnosis. We developed a benchmark dataset to evaluate and select best performing LLMs for medical reasoning.

    Therefore, to summarize the research insights, this study focuses on the following Research Questions (RQs):
    \begin{itemize}[leftmargin=10pt, itemindent=0pt, itemsep=0pt, topsep=7pt, parsep=0pt]
        \item \textbf{RQ1:} How does classification influence LLMs?
        \item \textbf{RQ2:} How effective are LLMs in clinical reasoning when DL is used as classifier?
        \item \textbf{RQ3:} How closely do LLM-generated rationales align with expert-annotated clinical reasoning?
    \end{itemize}

    Most of the existing studies either focus on classifying medical images without explaining the results or evaluate LLMs only through text-based tasks. As a result, there is still a gap between what is seen in the image and the kind of clinical reasoning needed to support real medical decisions. To bridge these gaps, our study introduces a unified framework that links image-based disease detection with LLM-driven reasoning and provides a systematic way to evaluate their combined performance. The notable contributions of this research are as follows:
    \begin{itemize}[leftmargin=10mm, itemindent=0mm, itemsep=0mm, topsep=5mm, parsep=0mm]
        \item We introduce a deep-learning–assisted LLM framework that produces richer and more reliable clinical reasoning than existing single-model approaches.
        \item We propose a structured model selection strategy that pairs deep learning models for medical image classification with appropriate LLMs for clinical reasoning.
        \item We develop MobileCoAtNet, a lightweight yet powerful hybrid deep learning architecture tailored for medical image classification. This combine multi-level convolutional feature fusion with transformer-based contextual modeling.
        \item We present the first open benchmarking platform designed to evaluate LLM reasoning quality driven by deep learning model outputs. The taxonomy organizes LLMs into three categories, high, moderate, and low clinical reasoning alignment-providing a transparent way to compare their strengths.
    \end{itemize}

    The remainder of this paper is organized as follows. \autoref{literature_review} reviews the existing literature relevant to this study. The problem definition and objectives are presented in \autoref{Problem_statement}. The proposed methodology, including data collection, feature extraction, and model design, is detailed in \autoref{methodology}. The experimental setup, including the dataset and implementation details, is described in \autoref{experimental_setup}. The results and key findings, along with comparisons to prior work, are reported in \autoref{Experimental Results}. A detailed discussion of the results is provided in \autoref{discussion}. The conclusions are drawn in \autoref{conclusion}, while the limitations of the study and possible future directions are outlined in \autoref{limitations}.

\section{Related Work}
\label{literature_review}
\subsection{Deep Learning based Classification}

Deep Learning (DL), particularly Convolutional Neural Networks (CNNs), has established a dominant position in medical image classification and analysis\cite{LI2024108635,FAHIM2025100571, AHMAD2024104528}. These models have significantly enhanced diagnostic precision by automating the detection of intricate visual features that are traditionally analyzed manually. However, a systematic review of the literature reveals that while these architectures excel at feature extraction and classification accuracy, they consistently operate as black box systems. They maintain a heavy reliance on statistical feature correlation rather than explicit diagnostic reasoning.

Early research concentrated on feature fusion and optimization within controlled or standard \cite{hasan2025dia,hasan2025transfusionnet}. Naz et al. \cite{naz2021segmentation} developed a dual-pipeline system combining image enhancement with parallel Inception V3 and LBP feature extraction. By selecting optimal features via Shannon Entropy for an Ensemble Learning Classifier, this method reached 99.80\% accuracy on a 3-class private dataset. Performance fell to 87.80\% on the public 8-class Kvasir dataset, indicating that generalization is restricted as class complexity grows. To improve efficiency in these standard environments, Khan et al. \cite{khan2025novel} proposed a network-level fusion of a Sparse Convolutional DenseNet201 with Self-Attention and a CNN-GRU hybrid. Their Shallow Wide Neural Network, optimized via an Entropy-controlled Marine Predators Algorithm, attained 99.60\% accuracy across 7 classes (Kvasir-V1) and 95.10\% across 8 classes (Kvasir-V2). Khan et al. \cite{khan2024deep} also applied a parallel strategy using Darknet-53 and Xception streams optimized by the Binary Dragonfly Algorithm. The concatenated vector was then classified with an Ensemble Subspace k-Nearest Neighbors classifier. This yielded 98.25\% accuracy on the 8-class Kvasir-V2, although the ESKNN classifier required longer training times.

Other studies aimed to improve the 8-class Kvasir benchmark by adding localization or attention mechanisms to generate more interpretable features. Alhajlah et al. \cite{alhajlah2023gastrointestinal} presented a multi-stage pipeline using Mask R-CNN to localize infections before feature extraction with fine-tuned ResNet-50 and ResNet-152 models. This resulted in 96.43\% accuracy, but the sequential steps added substantial computational overhead. Alternatively, Demirbaş et al. \cite{demirbacs2024spatial} proposed the Spatial-Attention ConvMixer, which integrated a Spatial Attention Mechanism to highlight specific regions. The model achieved 93.37\% accuracy on the 8-class Kvasir dataset. This was only a slight improvement over the baseline ConvMixer at 92.48\%, suggesting that attention mechanisms alone are insufficient for major performance gains without deeper reasoning capabilities.

Limitations in statistical DL approaches are evident when scaling to datasets with 23 or more classes. Rubab et al. \cite{rubab2025gastrointestinal} replaced pooling in ResNet architectures with an Entropic Field Propagation layer, optimized by a Newton-Raphson Marine Predator algorithm. Although it reached 99.0\% accuracy on the 8-class Kvasir V1, performance fell to 82.7\% on the complex 23-class Hyperkvasir dataset. Haseeb et al. \cite{haseeb2023fusion} applied a parallel-stream Nasnetmobile with a custom Stacked Auto-Encoder to the same 23-class dataset. The resulting 93.80\% accuracy proved slightly lower than the single-stream Nasnetmobile at 93.9\%, indicating the autoencoder added complexity without benefit. Ahamed et al. \cite{1} also designed a lightweight Parallel Depthwise Separable CNN optimized with the Pearson Correlation Coefficient. When applied to the 27-class GastroVision dataset, the model achieved 87.75\% accuracy. Authors attributed this result to severe class imbalance and the difficulty DL models face in distinguishing numerous classes without a robust reasoning framework.

While the reviewed literature confirms that DL models excel at image classification, they fundamentally lack reasoning capabilities. Across studies ranging from simple 8-class problems to complex 27-class datasets, no methodology successfully incorporated mechanisms to explain the logical basis of a diagnosis. The current state-of-the-art relies strictly on statistical pixel correlations, operating effectively as a black box. However, this dependence limits accuracy when scaling to high-dimensional data, as statistical features alone struggle to distinguish between numerous complex categories. This highlights a critical gap: the need to move beyond simple classification toward reasoning-based architectures that offer the logical transparency required for clinical practice.

\subsection{LLM-based Classification and Reasoning}

Recent advances in multi-modal LLMs have expanded their capabilities to integrate textual and visual medical data, enhancing diagnostic reasoning. Agbareia et al. \cite{agbareia2025visual} evaluated the visual-textual integration mechanism in GPT-4o and Claude Sonnet 3.5 against physicians on 120 clinical vignettes presented in text-only and text + image formats for medical diagnosis. Using a dataset from the OPENI database and recent NEJM challenges, LLMs outperformed physicians in text-only scenarios where GPT-4o achieved 70.8\% accuracy, Claude Sonnet 3.5 achieved 59.5\% accuracy, while physicians achieved 39.5\% accuracy. With the addition of images, all groups showed improved accuracy, but physicians demonstrated the largest accuracy gain (from 39.3\% to 78.8\%), significantly exceeding the LLMs' gains, where in the case of GPT-4o, accuracy increased from 13.7\% to 84.5\%; in Claude Sonnet 3.5, accuracy increased from 7.8\% to 67.3\%. Both models confirmed genuine integration of visual data by changing their explanations in 45-60\% of cases when images were provided. The core limitation identified is the LLMs' smaller relative gain, suggesting limited visual reasoning capacity, an open-evaluation process \cite{hasan2025pitfalls} and a continued heavy reliance on text-based knowledge compared to human clinicians. Similarly, Kaneyasu et al. \cite{KANEYASU2025} investigated the image-recognition capabilities of five multimodal LLMs on the Japanese National Examination for Dental Hygienists. The methodology used a zero-shot \cite{sarkar2025zero} evaluation on 213 multiple-choice questions (74 text-only, 139 visually-based). Gemini 2.5 Pro Experimental achieved the highest overall accuracy score of 85.0\%. Critically, for the visually-based questions, Gemini 2.5 was significantly superior at 82.0\% accuracy, outperforming other models, which averaged 70-73\% , suggesting that advanced image understanding is a new differentiator among LLMs. This highlighted progress in vision-language integration for clinical images. The principal limitation is the LLMs' imperfect accuracy (Gemini 2.5 still missed 15\% of the questions), resulting in an error rate too high for unsupervised use in high-stakes settings, compounded by potential linguistic and training-data biases \cite{das2024investigating} and the tendency toward hallucinations \cite{das2025investigating}. 

Rewthamrongsris et al. \cite{REWTHAMRONGSRIS2025100848} evaluated seven multimodal LLMs in the image-based diagnostic detection and differential diagnosis of Oral Lichen Planus (OLP), comparing them against specialized Convolutional Neural Networks (CNNs). The methodology tested the LLMs using a 1,142-image dataset across three designs: zero-shot recognition, example-guided recognition (a prompt engineering strategy), and differential diagnosis generation. Example-guided prompting improved performance; Gemini 1.5 Flash achieved the highest 80.53\% accuracy in the guided recognition task, while Claude 3.5 Sonnet excelled at ranking differential diagnoses. The core finding is a limitation: all LLMs were outperformed by the specialized CNN models for OLP detection. This discrepancy stems from LLMs being trained on publicly available data, which lacks sufficient domain-specific OLP-labeled examples, thereby limiting their single-task classification accuracy. The review asserts that LLMs possess powerful reasoning capabilities but are poor at classification. This suggests that when reasoning with image information, a combination of DL with LLM would be a better choice for improved reasoning in the medical imaging domain. 
\section{Problem Formulation}
\label{Problem_statement}


To investigate the influence of classification (e.g., classification of medical images) on LLMs (i.e., RQ1), we develop a fine-tuned hybrid deep learning model, $M^t$, for disease classification $C$ with high accuracy on the stomach disease dataset (Kvasir2) containing eight different classes.  
    \begin{equation}
    C = \Gamma(M^t, D)
    \end{equation}
Where $\Gamma()$ represents the function for classification. The classification information $C$ is then used in LLMs through prompt engineering to investigate its influence on reasoning. More details of the model $M$ are provided in Section 4.1.
\\
\\
To investigate the effectiveness of LLMs in classification and clinical reasoning (i.e., RQ2), we developed a two-stage evaluation framework to categorize LLMs based on classification and reasoning tasks. The benchmark dataset GastroLLM-Bench is used to evaluate 32 LLMs on a zero-shot learning task.  
    \begin{equation}
    \begin{split}
    LLM_j ={} & \psi(LLM_i, GastroLLM\text{-}Bench), \\
    & i = 1,2,\dots,32; j = 1,2,\dots,15
    \end{split}
    \end{equation}
    $\psi()$ represents the function for zero-shot learning. The 15 LLMs selected from zero-shot learning undergo two-step reasoning tasks: a multi-choice task and a reasoning task.  
\\
\\
The 15 selected LLMs are evaluated for a multi-choice task $T_m$ on the Gastro-reasonBench dataset that contains a set of multiple choice queries $Q_m$. The predicted label $L_p$ of the stomach image is included in the prompt.  
    \begin{equation}
    \begin{split}
    LLM_k = \Phi_1(LLM_i, T_m, Q_m, L_p), \\
    i = 1,2,\dots,15; k = 1,2,\dots,11
    \end{split}
    \end{equation}
    $\Phi_1()$ represents the function for a multi-choice task. 11 LLMs are selected based on the performance from the multi-choice task. 
\\
\\
The 11 models then undergo a reasoning task, $T_R$, which asks LLMs to generate justification for its own generated output, $O_m$, in a multiple-choice task, $T_m$. 
    \begin{equation}
    LLM_k = \Phi_2(LLM_k, T_R, Q_m, O_m, L_p), k = 1,2,\dots,11
    \end{equation}
    $\Phi_2()$ represents the function for the reasoning task. Further details of the multi-choice task and the reasoning task are provided in Section 4.2. 
\\
\\
To evaluate the reasoning alignment ability of the LLMs (i.e., RQ3), we developed a weighted score for each prompt and a final score (Context and Semantic alignment metrics) for rating the LLMs as High-CRA, Moderate-CRA, or Low-CRA.  
    \begin{equation}
    ^\lambda{LLM}_k = \Omega[\Phi_2(LLM_k, \text{FinalScore})]
    \end{equation}
    $\Omega[]$ represents the evaluation function for human-like reasoning and $\lambda \in \{\text{High-CRA, Moderate-CRA, Low-CRA}\}$. The 11 LLMs are categorised according to their human-level reasoning performance. The categories are \{High-CRA, Moderate-CRA, Low-CRA\}. Further details of the categorisation of LLMs are provided in Section 4.3.

\begin{figure*}[p]
    \centering
    \begin{subfigure}[b]{\textwidth}
        \centering
        \includegraphics[width=0.85\linewidth]{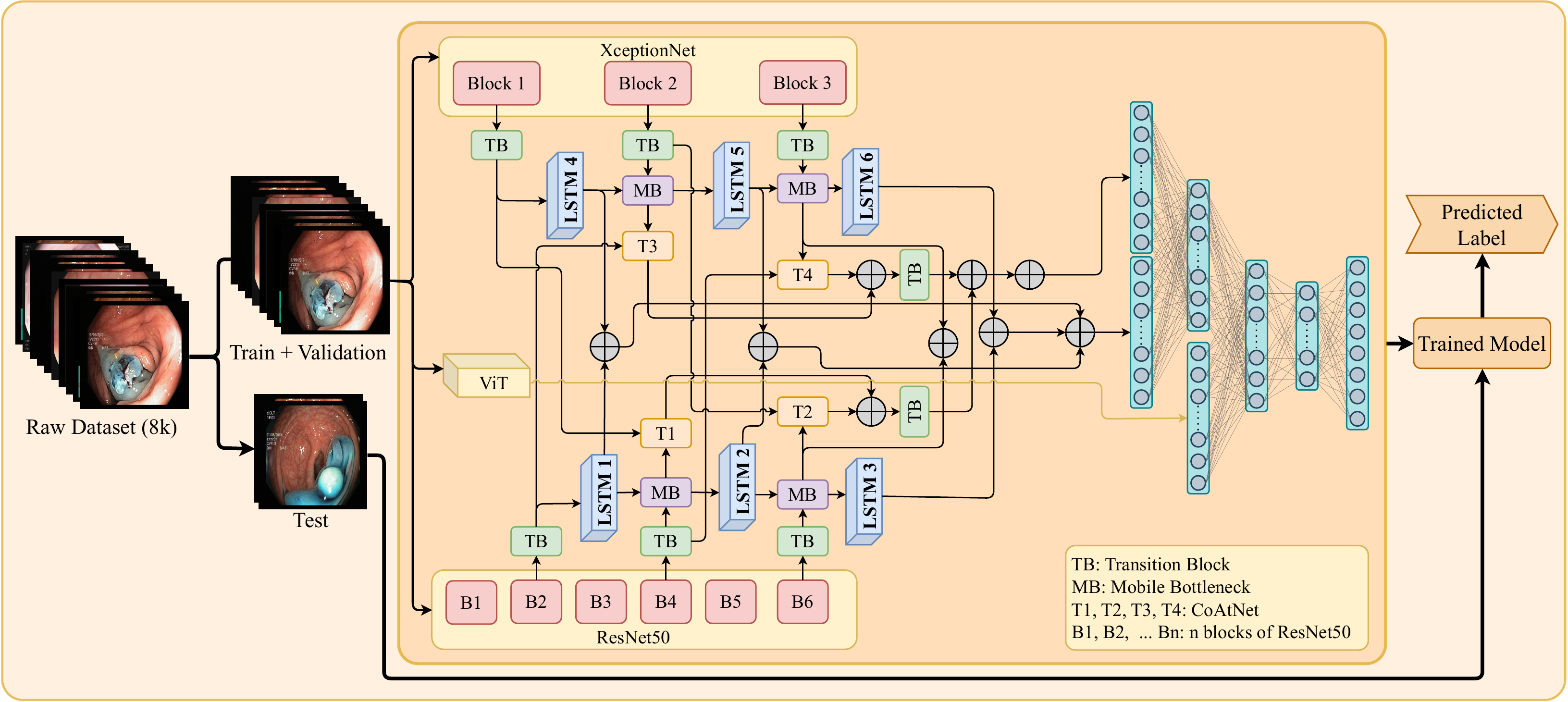}
        \caption{The working principle of MobileCoATNet.}
        \label{fig:method_a}
    \end{subfigure}
    
    \vspace{0.5em} 

    \begin{subfigure}[b]{\textwidth}
        \centering
        \includegraphics[width=0.75\linewidth]{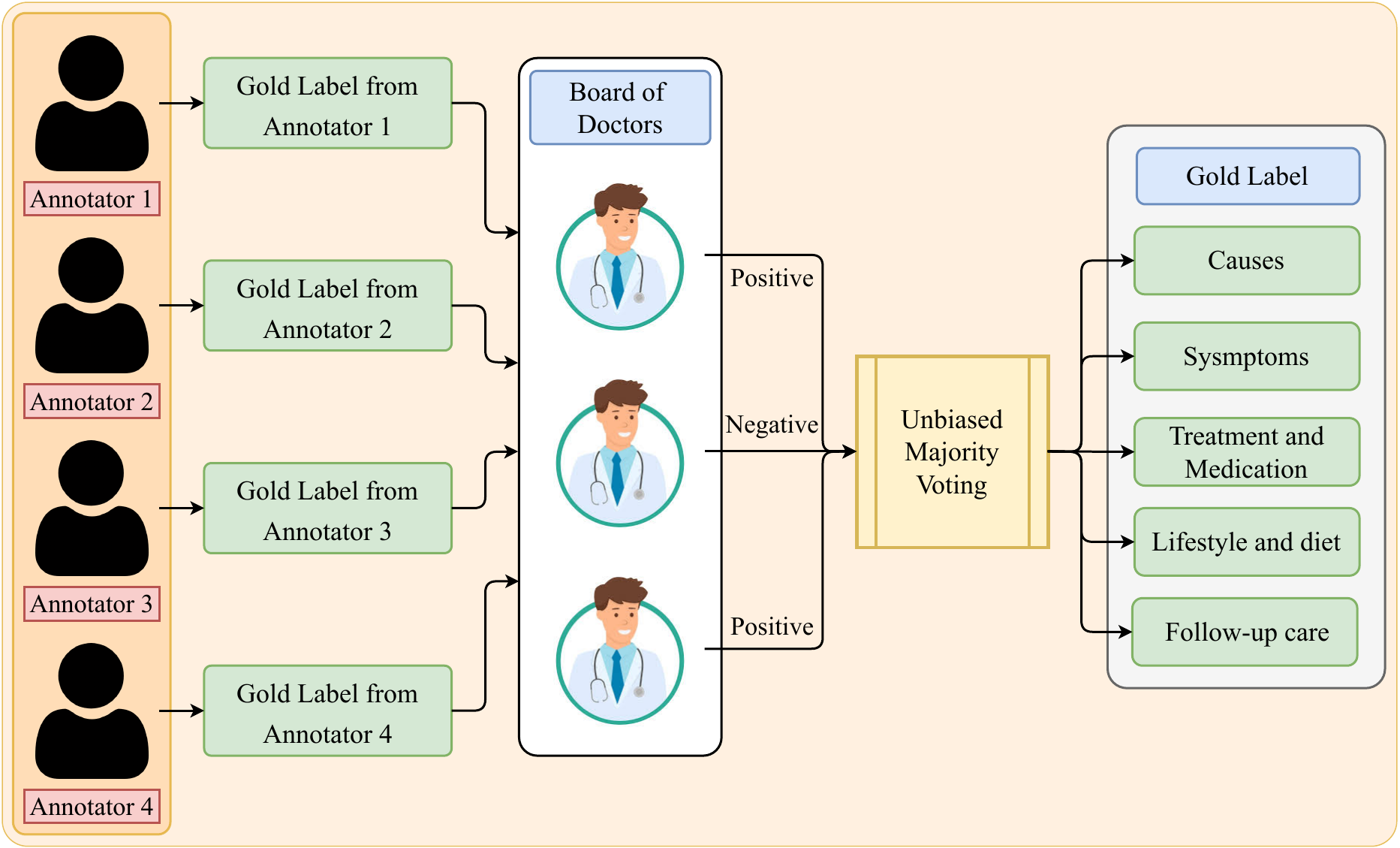} 
        \caption{A framework to generate human annotation for the benchmarks.}
        \label{fig:method_b}
    \end{subfigure}

    \vspace{0.5em}

    \begin{subfigure}[b]{\textwidth}
        \centering
        \includegraphics[width=0.85\linewidth]{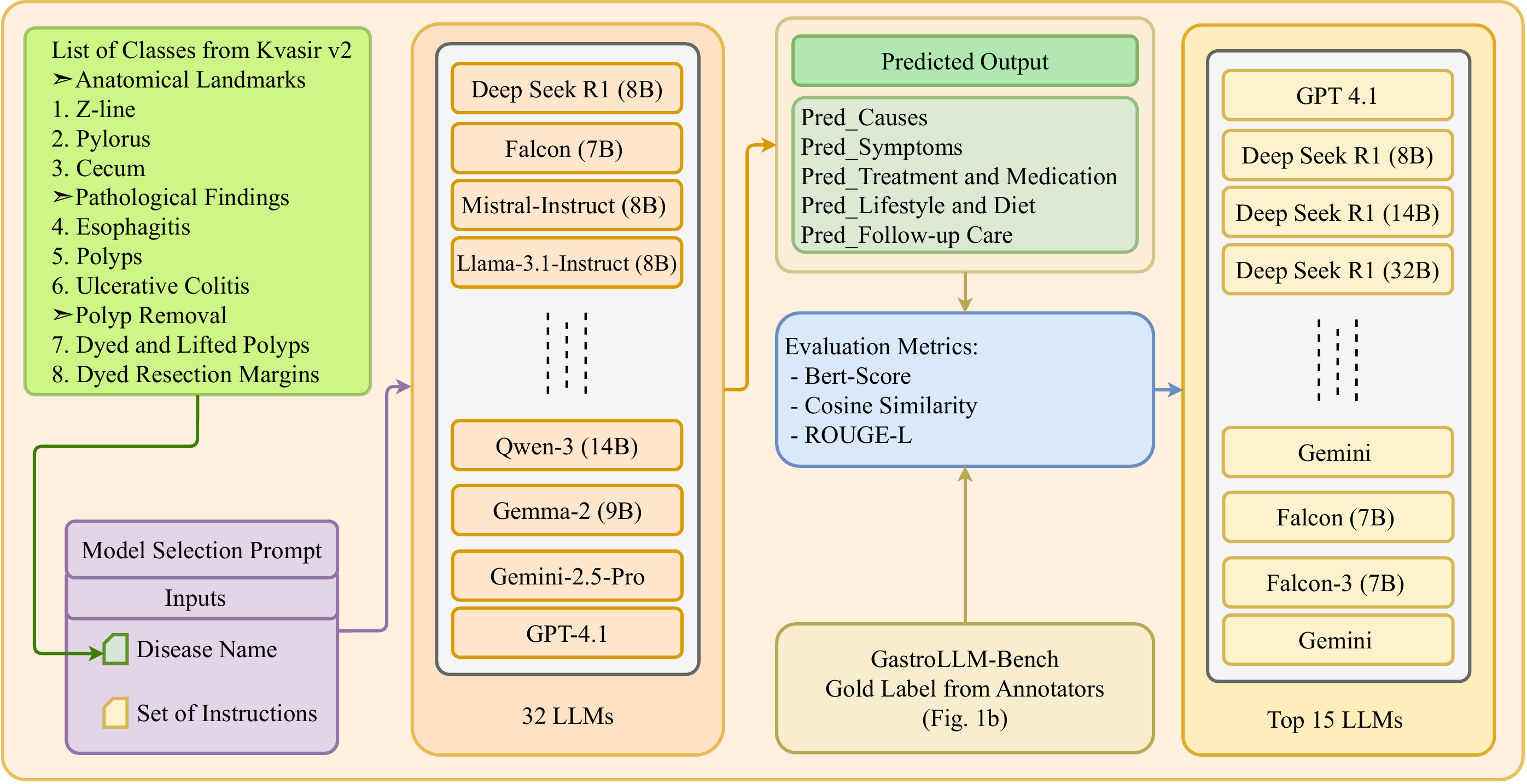}
        \caption{Benchmarking for initial LLM selection.}
        \label{fig:method_c}
    \end{subfigure}

    \label{fig:overall_methodology_split}
\end{figure*}

\begin{figure*}[t] 
    \ContinuedFloat 
    \centering
    \begin{subfigure}[b]{\textwidth}
        \centering
        \includegraphics[width=\linewidth]{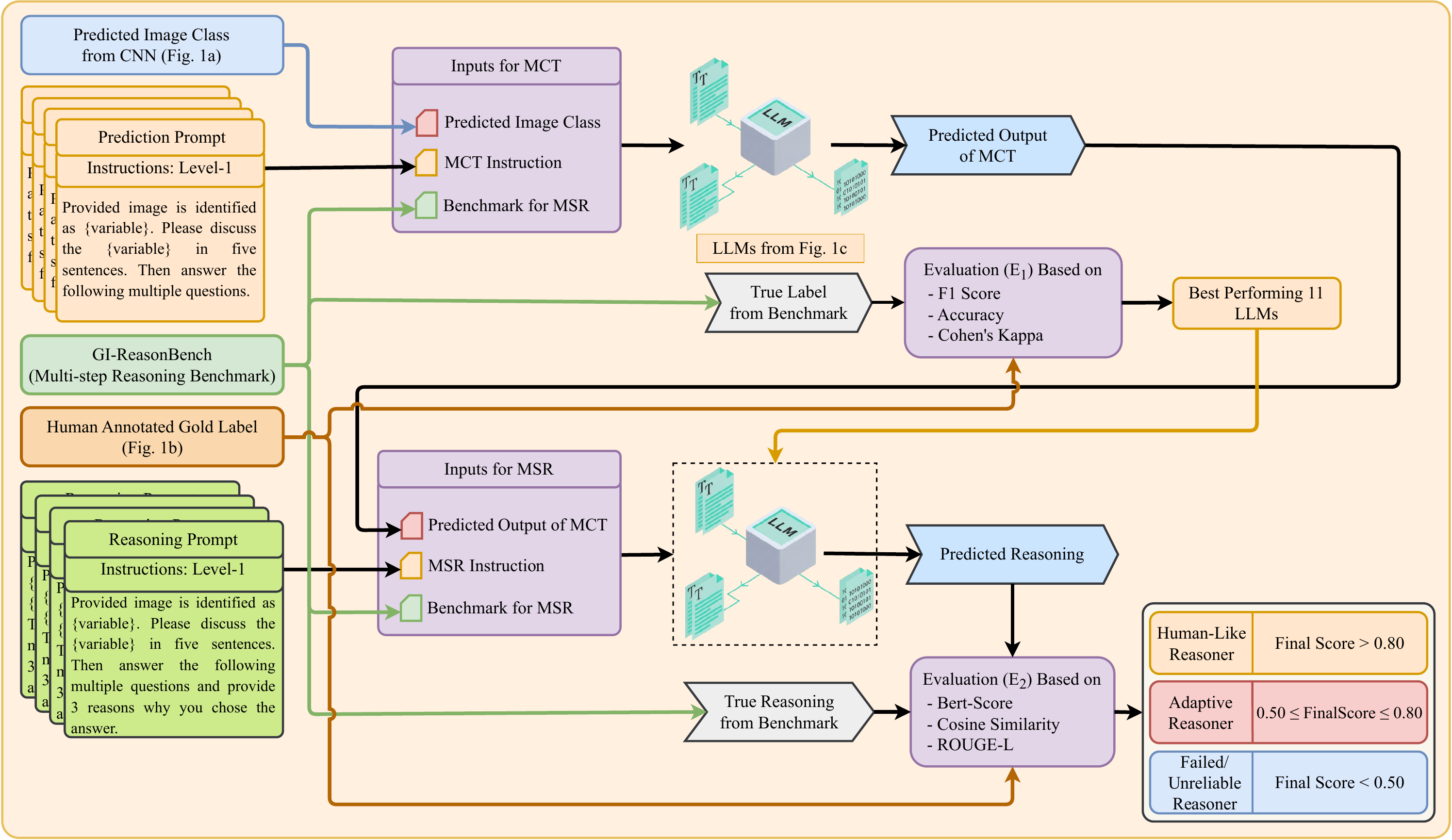}
        \caption{DL-based multi-step reasoning using LLMs.}
        \label{fig:method_d}
    \end{subfigure}

    \caption{Overall methodology: (a) Medical image classification using MobileCoAtNet to diagnose the disease level, (b) A multi-step majority voting-based benchmark annotation pipeline for gold-label generation, (c) A pipeline for zero-shot-based initial LLM selection for clinical reasoning tasks, and (d) The image classified from a DL model and the predicted label is sent to an LLM for multi-step clinical reasoning generation.}
    \label{fig:overall_methodology_split_cont}
\end{figure*}
\noindent

\section{Methodology}
\label{methodology}
LLMs have been widely used in medical diagnosis and reasoning with great success. Very often medical images, e.g. stomach infection endoscopy images, are used to enhance reasoning in LLMs. LLMs are inherently designed for text processing and understanding but they are not efficient in classification of images. Therefore, prior classification of images using DL models and integrating it with LLMs may prove to be useful. The integration of DL models for classification and LLMs for medical reasoning is non-trivial due to their unique characteristics, which requires an evaluation and selection strategy for LLMs based on their reasoning capability. LLMs exhibit impressive reasoning performance on many benchmarks, but it is not comparable to that of human. Therefore, the reasoning performance of LLMs is to be evaluated against benchmark using some metrics. The methodology comprises three sections: classification using DL models (section 4.1), reasoning using LLMs (section 4.2), and evaluation of reasoning performance of LLMs (section 4.3). The methodology showing the three sections is presented in Figure 2.   

\subsection{Classification using Deep Learning}
Classification medical images is performed using DL models prior to their use in LLM. In order to find the best performing DL model, a set of 26 models (see Table 3) was chosen for investigation. They are evaluated using zero-shot task on stomach infection endoscopy images. A subset of eight DL models is selected through the zero-shot task based on accuracy and F1-score. The eight pretrained models and a proposed model called MobileCoAtNet are then fine-tuned and evaluated using five-fold cross-validation where the performance is assessed on a held-out test set based on test-accuracy, F1-Score, and Cohen’s Kappa (see Table 4). 

The architecture of MobileCoAtNet (see \autoref{fig:method_a}) comprises two branches: an upper branch and a lower branch. These branches can incorporate various combinations of base models to construct the complete proposed framework. Three base models were considered: ViT, ResNet50, and XceptionNet based on their performance. In the final configuration, the upper branch employs XceptionNet, while the lower branch utilizes ResNet50.

Preprocessed images with a size of \(128 \times 128 \times 3\) were fed into pretrained XceptionNet, ResNet50, and ViT models, which were utilized without their top layers to preserve general features while enabling domain adaptation for endoscopic analysis. From each branch, three layers were selected to extract features that were then passed through separate transition\_block modules using a consistent filter size of 512. Structurally, each transition\_block consisted of a loop containing a separable convolution layer that is activated only when the first dimension is greater than 4.

Subsequently, outputs from six transition blocks with a shape of \(4 \times 4 \times 512\) were processed by six LSTMs to produce vectors of size 1024. These outputs were reshaped into \(4 \times 4 \times 256\) and concatenated with the corresponding transition\_block features to result in a combined shape of \(4 \times 4 \times 768\).

The combined features from each branch were passed into the \text{mobile\_bottleneck} producing an output with a dimension \(4 \times 4 \times 512\), which is repeated for all four feature sets. The outputs of each branch’s \text{mobile\_bottleneck} were concatenated with the corresponding transition\_block outputs from the opposite branch, producing a fused feature map of \(4 \times 4 \times 1024\).

The fused features were processed through four \text{CoATNet} modules. In each branch, outputs from two modules were concatenated and refined via a \text{transition\_block} to form a \(4 \times 4 \times 512\) feature map. The resulting \text{transition\_block} outputs were then merged to produce a final \(4 \times 4 \times 1024\) feature map.

The aggregated output was fused with the concatenated features from the final \text{mobile\_bottleneck} of each branch, producing a \(4 \times 4 \times 2048\) feature map. This was flattened into \(A\) with 32,768 elements, while the outputs of all six LSTM units were combined into \(B\) with 6,144 features.

Vectors \(A\) and \(B\) were concatenated to form \(C\) with 38,912 features. A sequence of \text{Dense} and \text{Dropout} layers reduced \(C\) to a vector \(D\) of 2,048 features. \(D\) was then merged with the ViT output vector \(E\) (1,000 features) to produce the final combined vector \(F\) with 3,048 features. Multiple Dense and \text{Dropout} layers were applied to \(F\) to generate the predicted output. 

For fine-tuning the DL models, the image dataset is divided into training, validation, and testing subsets with a ratio of 70:20:10. During the fine-tuning process, all pretrained layers of the selected models are frozen to preserve their learned representations, while only the output layers are trained to adapt to the new task.

The classification information from MobileCoatNet, i.e. predicted labels, is then fed into LLMs through prompt engineering to assess the reasoning ability of the models.

\begin{figure*}[t!]
    \centering
    \begin{subfigure}{0.85\linewidth}
        \centering
        \includegraphics[width=\linewidth]{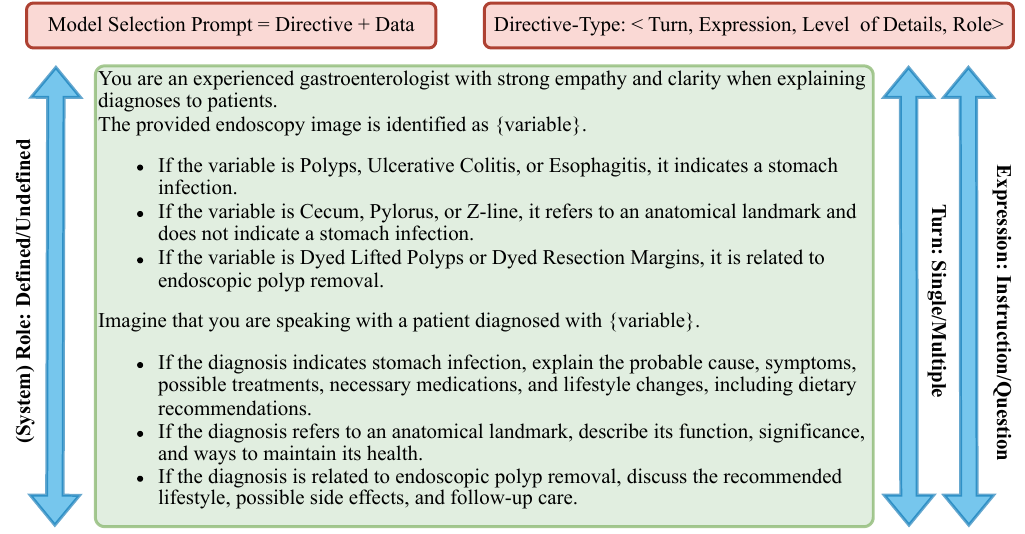}
        \caption{Prompt for zero-shot-based initial LLM selection.}
        \label{fig:Model Selection Prompt}
    \end{subfigure}
    
    \vspace{0.2cm}
    
    \begin{subfigure}{0.9\linewidth}
        \centering
        \includegraphics[width=\linewidth]{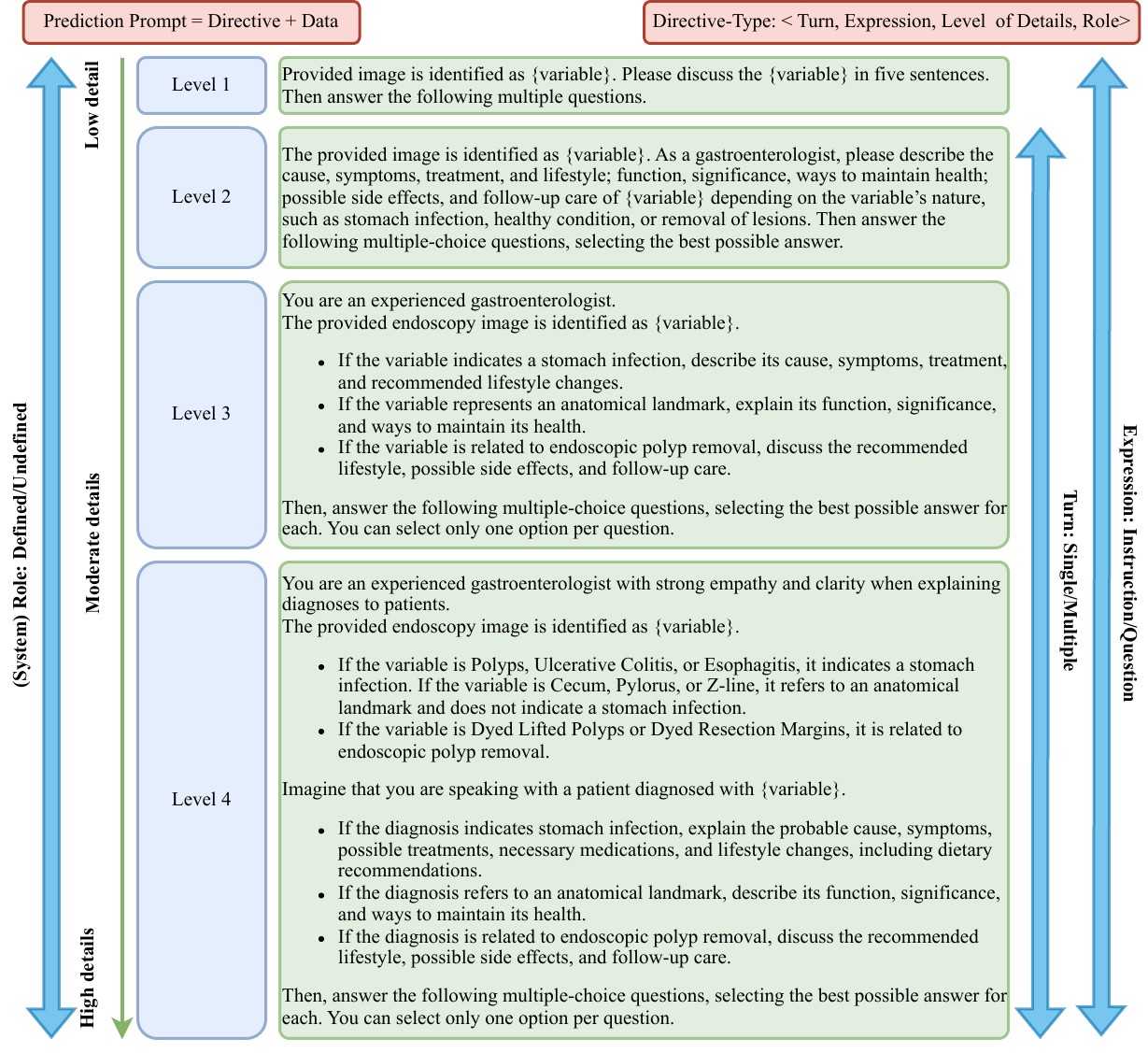}
        \caption{Prompt for LLM-based multiple-choice task.}
        \label{fig:Prediction Prompt}
    \end{subfigure}
\end{figure*}

\begin{figure*}[t!]
    \ContinuedFloat
    \centering

    \begin{subfigure}{0.9\linewidth}
        \centering
        \includegraphics[width=\linewidth]{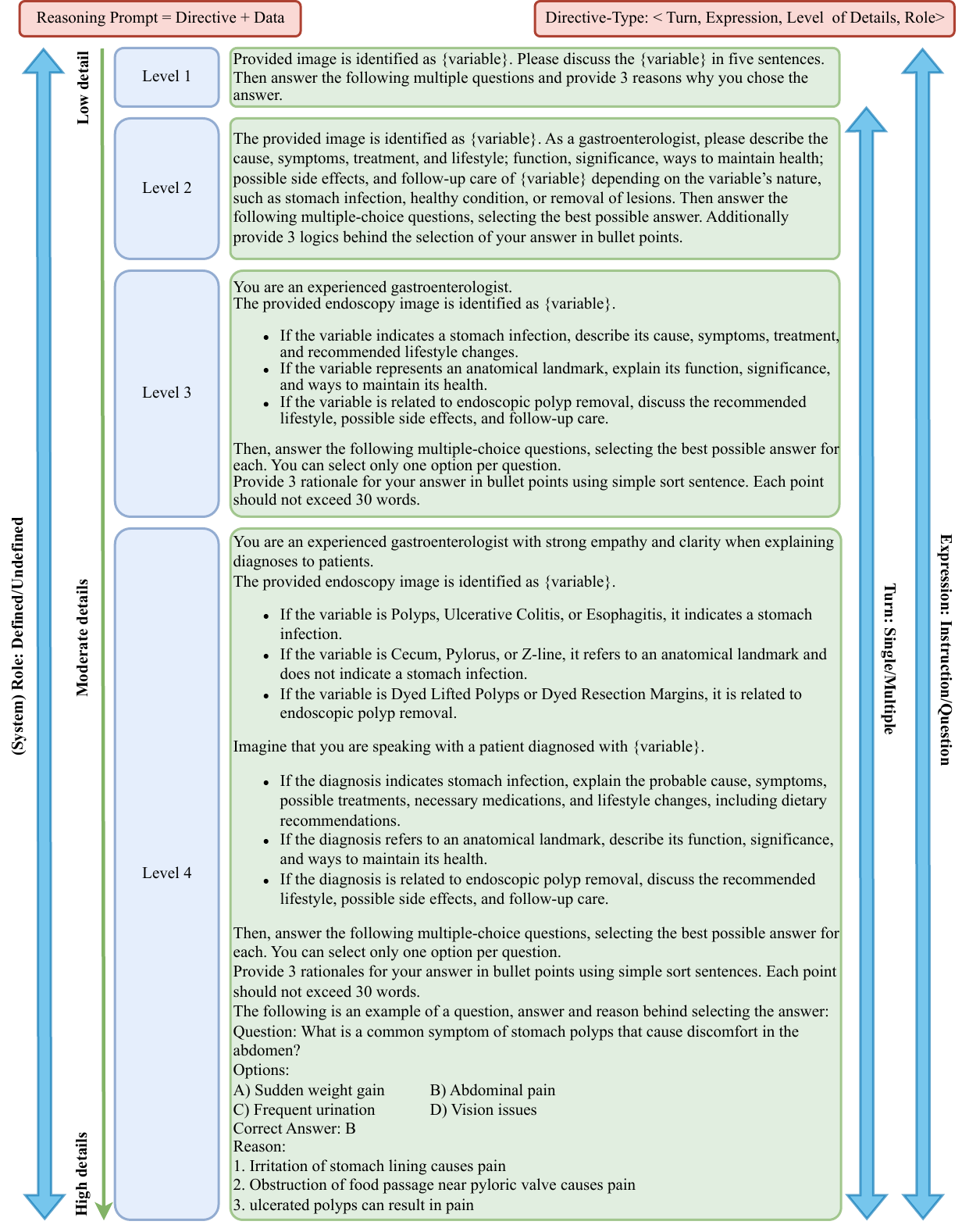}
        \caption{Prompt for LLM-based reasoning task.}
        \label{fig:Reasoning Prompt}
    \end{subfigure}
    
    \caption{Three different prompts, (a), (b), and (c), have been used for three different tasks. For each task, we have used a taxonomy of prompts, where four different levels of prompts are used with varying levels of directives.}
    \label{fig:all_prompts}
\end{figure*}

\subsection{Reasoning using LLM}
As the objective is to evaluate LLMs for reasoning and assess their generalizability in medical imaging domain, we will select a range of models from the most used LLM families such as OpenAI, Gemini, LLaMA, DeepSeek, Qwen, and Falcon based on three different categories of parameter size - (i) below 10 billion parameters, (ii) 10 to 50 billion parameters, and (iii) more than 50 billion parameters. For this experiment, we chose a total of 32 LLMs (see Table 6), of which 19 are open-source and 13 are proprietary. 

The 32 LLMs are evaluated on GastroLLM-Bench using zero-shot prompting strategy (see \autoref{fig:Model Selection Prompt}) for selection of LLMs that guides LLMs in generating relevant responses from the Kvasir-2 dataset. As shown in Figure 3a, the prompt provides each class name together with a set of queries. This ensures that every model consistently interprets the dataset’s categories such as anatomical landmarks, stomach infection, and endoscopic polyp removal and generates responses aligned with clinical reasoning. Each model produced outputs for the defined categories, and these responses are compared against human annotated gold-labels. Three evaluation metrics—BERTScore, Cosine Similarity, and ROUGE-L are used to assess semantic accuracy, contextual relevance, and textual coherence. Based on the three evaluation metrics, top-performing 15 models are selected for multi-step reasoning performance. 

To investigate LLMs’ knowledge depth and proper generalization, a multi-step reasoning evaluation framework is developed for two different tasks - Multiple-choice task and Reasoning task.

(A) Multiple-choice Task: The selected models are evaluated using four prompt sets, including zero-shot and one-shot variants (Figure 3b). In this stage, each LLM
receive images that have been classified by the DL model, along with their predicted labels. Based on the predicted class, the LLMs are prompted to provide specific
information. For stomach infection classes, they are given multiple-choice queries covering causes, symptoms, treatment options, and lifestyle recommendations, and are asked to select the most suitable answer. For anatomical landmark classes, the queries focused on function, clinical significance, and strategies for maintaining overall health. For endoscopic polyp removal classes, the questions address lifestyle
guidance, potential side effects, and recommended follow-up care. The outputs produced by the LLMs are compared with gold labels and evaluated using Accuracy, F1-score, and
Cohen’s Kappa. Based on these results, the best-performing 11 models are selected for the reasoning task and further analysis.
   
(B) Reasoning Task: The top-performing 11 models from the multiple-choice task stage are evaluated on reasoning task. In this experiment, each LLM receive several components
through the prompt: a question-answer pair where the answer are generated by that same LLM, an endoscopic image with its predicted label, and an instruction to provide
three justifications explaining why the predicted answer aligns with the given question. The full prompts are shown in Figure 3c. The generated reasoning is compared with
human-produced gold reasoning using BERTScore, cosine similarity, and ROUGE-L to identify the best-performing models.

The two-stage evaluation framework is established to categorize LLMs on Gastro-reasonBench based on their reasoning performance and classification tasks. 

\textbf{Benchmark Design:} The benchmark design begins with the development of two comprehensive “Gold Label” benchmarks focused on the gastrointestinal tract: one for model selection and the other for multi-step reasoning. The process begins with an extensive review of research papers, official health organization websites, and other credible medical sources to ensure accurate initial annotations. For LLM evaluation, the original eight classes were consolidated into five to align with common clinical decision categories. The three anatomical landmarks (Z line, Pylorus, and Cecum) are merged into the Healthy Class, while the two polyp removal categories (Dyed and Lifted Polyps and Dyed Resection Margins) are combined as Removal of Lesions. The remaining pathological classes (Esophagitis, Polyps, and Ulcerative Colitis) are retained as separate labels. The final label set consists of Healthy, Esophagitis, Polyps, Ulcerative Colitis, and Removal of Lesions.

Detailed annotations are created under five medical categories: `Causes’, `Symptoms’, `Treatment’ and Medication’, `Lifestyle and Diet’, and `Follow-up Care’. These annotations are independently prepared by five human experts, identified as Gold Label from ``Annotator 1” through ``Annotator 5.” The annotated data are then reviewed and refined through majority voting with a ``Board of Doctors” to ensure medical-level accuracy and consistency, resulting in the final version of the ``Gold Label” benchmarks.

GastroLLM-Bench consists of information to evaluate 32 LLM models on a zero-shot prompt (\autoref{fig:Model Selection Prompt}). The information gathered consists of all possible causes, symptoms, treatment and medication, lifestyle and diet, and follow-up care. Gastro-reasonBench is created with multiple-choice questions across the 5 categories (Causes, Symptoms, Treatment and Medication, Lifestyle and Diet, and Follow-up Care) with reasons behind the right answers. The Gastro-ReasonBench consists of 375 questions, 100 questions each for the disease classes: Polyps, Ulcerative Colitis, and Esophagitis. Additionally, there are 25 questions related to healthy anatomical classes such as the Cecum, Pylorus, and Z-line. These questions focus on the importance of these anatomical structures and lifestyle practices for maintaining a healthy stomach. Lastly, 50 questions are designed for the removal of lesions, including Dyed Lifted Polyps and Dyed Resection Margins, with questions regarding their purpose, potential side effects, medical follow-up, and associated lifestyle and dietary habits.

\subsection{Evaluation Strategy of LLM reasoning}
A systematic evaluation of LLM reasoning is carried out to measure their effectiveness in handling disease-related queries in the gastrointestinal domain. 

\textbf{Prompting Strategies:} To investigate prompt sensitivity of LLMs, we developed four sets of prompts for each benchmark following the established TELeR-taxonomy of prompts ~\cite{karmaker2023teler}. Prompts one, two, and three are zero-shot prompts, while prompt four is one-shot prompt. We structured the prompting sequence progressively, starting from general instructions and moving toward more specific guidance, with the final prompt including an illustrative example. \autoref{fig:Prediction Prompt} presents an example from the benchmark.

To assess the effectiveness of the proposed framework, a multi-stage evaluation protocol is established for both the deep learning visual classifiers and the large language reasoning models. The mathematical formulations for the quantitative metrics utilized throughout this study are summarized in Table \ref{Performance metrics of multi-class classification}.

\textbf{Metrics for LLM Reasoning Performance:}
    To evaluate the models' ability to perform reasoning with minimal context, ``Final Weighted Score'' is computed, giving greater importance to prompts containing less specific information. For each prompt, a Composite Reasoning Score (CRS) is first calculated by averaging across all five classes and applying weighted metric importance.


\begin{equation}
\begin{split}
\text{CRS}_{\text{p}} 
&= w_1 \cdot \text{Avg\_Cosine}_{p}
 + w_2 \cdot \text{Avg\_BERT}_{p} \\
&\quad + w_3 \cdot \text{Avg\_Rouge}_{p}
\end{split}
\label{eq:crs_prompt}
\end{equation}

 To determine the weights {w1, w2, w3}, an ablation study is carried out on Gemini-1.5, Gemini-1.5-8B, Gemini-1.5-Pro, Gemini-2.0, GPT-3.5, GPT-4.1, GPT-4.1-mini, and GPT-4.1-nano (\autoref{tab:crs_prompt_weights}). Cosine similarity is given the highest weight ($w_1$) to emphasize global semantic alignment and the overall direction of the reasoning. It operates on aggregated sentence embeddings, and it is particularly effective at capturing whether the generated rationale matches the underlying meaning of the gold reasoning, even when the wording differs. In contrast, BERTScore offers useful insight into token-level contextual accuracy but is weighted lower ($w_2$) because its token-wise alignment can penalize valid reasoning that uses different phrasing or vocabulary to express the same idea. Finally, ROUGE-L is included with the smallest weight ($w_3$), serving mainly to measure exact textual overlap.
    
    The \emph{Final Weighted Score} for each model is then obtained as a weighted sum of the prompt-specific CRS values:
    
    \begin{equation}
    \text{FinalScore} = \sum_{i=1}^{4} w_i \, \text{CRS}_i 
    \end{equation}
    
    We assign decreasing weights (0.4, 0.3, 0.2, 0.1) to the four levels of prompt (see \autoref{fig:Reasoning Prompt}) because the earlier prompts (P1, P2) are designed to capture a model’s general reasoning stability, while the later prompts (P3 , P4) probe more targeted or assistive reasoning under additional guidance. As a result, P1 carries the highest weight (0.4) because it reflects the model’s “raw” zero-shot capability without external scaffolding. P2 (0.3) evaluates slightly aided reasoning but still preserves open-ended model behavior. P3 (0.2) and P4 (0.1) contribute less to the final score because they increasingly rely on prompt cues and therefore represent less of the model's intrinsic reasoning ability. This descending weighting follows prior evaluation practice in LLM robustness studies, where unguided reasoning is considered the most informative indicator of general model performance.

    The thresholds are used to categorize LLMs according to their level of \emph{Clinical Reasoning Alignment (CRA)} with the expert-annotated benchmark. Models with $\text{FinalScore} > 0.80$ are classified as \emph{High-CRA}, indicating strong semantic and contextual alignment with the reference reasoning across different prompt configurations, together with relatively low sensitivity to prompt variation. However, this category should not be interpreted as evidence of human-equivalent reasoning or readiness for autonomous clinical deployment. Human supervision remains necessary, particularly in high-stakes medical settings where even minor reasoning errors may have significant consequences. Models with $0.50 \leq \text{FinalScore} \leq 0.80$ are classified as \emph{Moderate-CRA}. These models demonstrate partial alignment with the expert-annotated reasoning and generally produce clinically relevant responses, but their performance may vary across prompt formulations and levels of contextual guidance. Although such models show promising reasoning capability, additional domain adaptation, fine-tuning, or task-specific optimization may be required before they can be considered reliable for specialized downstream applications. Models with $\text{FinalScore} < 0.50$ are classified as \emph{Low-CRA}. These models exhibit comparatively limited alignment with the expert-annotated reasoning and greater variability across prompting conditions. Their generated rationales may show weaker semantic consistency, reduced contextual agreement, or higher sensitivity to changes in prompt formulation. Consequently, these models require substantial improvement through domain-specific training, instruction tuning, or carefully curated clinical datasets before they can be considered suitable for practical medical reasoning applications. The proposed CRA categories therefore represent benchmark-relative levels of reasoning alignment rather than direct measures of human-like intelligence, clinical competence, or deployment readiness.

\begin{table}[t]
\small
    \caption{The set of performance metrics used to evaluate DL-based classification and LLM-based multi-step reasoning generation.}
    \label{Performance metrics of multi-class classification}
    \setlength{\tabcolsep}{2pt}
    \renewcommand{\arraystretch}{1.8}
    \begin{tabularx}{\linewidth}{p{2.2cm}r}
    \toprule
    \textbf{Metric} & \textbf{Formula} \\ 
    \midrule
    Accuracy  &
    $\text{ACC}= \frac{\text{TP} + \text{TN}}{\text{TP} + \text{TN} + \text{FP} + \text{FN}}$ \\
    Loss &
    $L = \text{Loss}(y,\hat{y})$ \\
    Precision &
    $\text{PPV}= \frac{\sum_{k} \text{TP}_{k}}{\sum_{k} (\text{TP}_{k} + \text{FP}_{k})}$ \\
    Sensitivity &
    $\text{SN}= \frac{\sum_{k} \text{TP}_{k}}{\sum_{k} (\text{TP}_{k} + \text{FN}_{k})}$ \\
    Specificity &
    $\text{SP}= \frac{\sum_{k} \text{TN}_{k}}{\sum_{k} (\text{TN}_{k} + \text{FP}_{k})}$ \\
    F1-Score &
    $F_{1} = \frac{2 \times \text{PPV} \times \text{SN}}{\text{PPV} + \text{SN}}$ \\
    Cohen's Kappa & 
    $\kappa = \frac{P_o - P_e}{1 - P_e}$ \\ 
    & $P_o = \text{ACC}$ \\ 
    & $P_e = \sum_{k=1}^{n} \frac{(\text{TP}_k + \text{FP}_k) \cdot (\text{TP}_k + \text{FN}_k)}{(\text{TP}_k + \text{TN}_k + \text{FP}_k + \text{FN}_k)^2}$ \\
    
    BERT-Score &
    $\text{BS} = \frac{1}{N} \sum_{i=1}^{N} 
    \max_{j}\; \text{sim}(e^{ref}_{i},\, e^{pred}_{j})$ \\
    ROUGE-L &
    $\text{ROUGE-L} = 
    \frac{2 \times \text{LCS}(y,\hat{y})}{|y| + |\hat{y}|}$ \\
    Cosine Similarity &
    $\text{CosSim} = 
    \frac{\vec{y} \cdot \vec{\hat{y}}}{\|\vec{y}\|\,\|\vec{\hat{y}}\|}$ \\
    \bottomrule
    \end{tabularx}
\end{table}
\section{Experimental Setup}
\label{experimental_setup}
\subsection{Dataset Description}
    In this study, the training and evaluation of DL models were performed using the Kvasir-V2 dataset \cite{kvasir}. It is a well-curated collection of 8,000 high-resolution gastrointestinal endoscopy images specifically designed for research and benchmarking, balanced equally across eight distinct classes. This dataset, originating from Vestre Viken Health Trust in Norway, encompasses three anatomical landmarks (Z-line, Pylorus, and Cecum), three pathological findings (Esophagitis, Polyps, and Ulcerative Colitis), and two classes related to polyp removal procedures (Dyed and Lifted Polyps and Dyed Resection Margins). All images, ranging in resolution from 720 $\times$ 576 to 1920 $\times$ 1072 pixels, were captured using advanced endoscopic equipment and subsequently verified by experienced gastroenterologists to ensure high clinical accuracy, and ground truth reliability. The dataset provides a robust foundation for developing and benchmarking automated systems for gastrointestinal disease analysis.

\subsection{Deep Learning Models}
    MobileCoAtNet comprises three main pretrained models XceptionNet \cite{xceptionnet}, ResNet50 \cite{resnet50}, and ViT \cite{vit_main}. Also the bottleneck of a lightweight model, MobileNetV2, and LSTM blocks/units are used in this study.

    \subsection{Large Language Models}
    Thirty-two LLMs from six families were evaluated: OpenAI (GPT-4, GPT-3.5), Gemini (Gemini 1, Gemini 1.5), LLaMA (LLaMA 2-7B), DeepSeek (DeepSeek R1), Qwen (Qwen-7B), and Falcon (Falcon-7B, Falcon-40B).  On the other hand, closed source models are used since they are trained on huge parameters with improved performance \cite{10.1007/978-3-031-53085-2_20}.

\subsection{Hyperparameters of DL models and LLMs}

\begin{table}[t]
    \centering
    \small
    \caption{Hyperparameters used for DL-based classification and LLM-based multi-step reasoning generation.}
    \label{tab:hyperparameters}
    \setlength{\tabcolsep}{4pt}
    \renewcommand{\arraystretch}{0.9}
    
    \begin{tabularx}{\linewidth}{llr}
    \toprule
    \textbf{Category} & \textbf{Hyperparameter} & \textbf{Value} \\
    \midrule
    
    \multirow{9}{*}{\makecell[l]{DL\\(MobileCoAtNet)}}
    & Input size & $128 \times 128 \times 3$ \\
    & Patience & 20 \\
    & Batch size & 128 \\
    & Metrics & Accuracy \\
    & Early stopping & True \\
    & Loss & \makecell[r]{Sparse categorical\\cross-entropy} \\
    & Monitor & Val\_accuracy \\
    & Optimizer & Adamax \\
    & Mode & Max \\
    \midrule
    
    \multirow{5}{*}{\makecell[l]{Open-source\\LLMs}}
    & Temperature & 0.7 \\
    & Max tokens & 50 \\
    & Top-$p$ & 1.0 \\
    & Do sample & True \\
    & Context length & $\sim$32K \\
    \midrule
    
    \multirow{3}{*}{\makecell[l]{Closed-source\\LLMs}}
    & Temperature & 0.7 \\
    & Max tokens & 50 \\
    & Context length & 128K \\
    \bottomrule
    \end{tabularx}
\end{table}

\label{Hyperparameters used in deep learning algorithms}
    This section outlines the specific hyperparameters utilized for training and inference across the different DL models and LLMs. These settings, summarized in \autoref{tab:hyperparameters}, were systematically chosen to ensure optimal performance, efficient training dynamics, and reproducibility of the results. 

    \textbf{(a) DL Models: }The MobileCoAtNet model was trained using $128 \times 128 \times 3$ images and a batch size of 128. The Adamax optimizer was chosen, with accuracy selected as the performance metric. The `sparse categorical cross-entropy' loss function was used, as it is suitable for multi-class tasks with integer labels. The loss for a single sample $i$ is defined as:
    \begin{equation}
        L(y, \hat{y}) = -\log(\hat{y}_{i, y_i})
    \end{equation}
    Here, $y_i$ is the true class index and $\hat{y}_{i, y_i}$ is the predicted probability for that class. To prevent overfitting, early stopping was applied. This monitored the validation accuracy in max mode and stopped the training if no improvement occurred for 20 consecutive epochs.

    \textbf{(b) LLMs: }The hyperparameters for the LLMs were configured to optimize their generative performance for the diagnostic tasks, with slight variations between open and closed-source models.

\begin{itemize}
    \item \textbf{Open-source Models}
    For the open-source LLMs, a probabilistic sampling approach was enabled. The generation was controlled by a temperature of 0.7 to balance creativity and coherence, while top p was set to 1.0. These models operated with a context length of approximately 32K tokens ($\sim$32K) and were limited to a Max token output of 50 new tokens.

    \item \textbf{Closed-source Models}
    The closed-source models were similarly configured with a temperature of 0.7 and a Max token limit of 50. These models, however, leveraged a significantly larger Context length of 128K tokens, allowing for the processing of more extensive input prompts.
\end{itemize}


    \subsection{Environment Setup}
    Both cloud and personal computing resources were used to develop and test the proposed hybrid deep-learning model for identifying stomach infection from gastrointestinal endoscopy images. The initial development took place on Kaggle, which provided each account with 15 GB of RAM, 170 GB of storage, and 30 GB of GPU memory. Preliminary model testing was conducted on Google Colab, where each account had 15 GB of GPU memory, 13 GB of RAM, and 10 GB of local storage. Additionally, the dataset was preprocessed on a personal computer with a Core i7 CPU, 32 GB of RAM, and an NVIDIA GPU.
    
\begin{table*}[t]
    \centering
    \small
    \caption{Zero-shot medical image (endoscopic image of stomach disease) classification using DL models.}
    \label{tab:model_performance}
    \setlength{\tabcolsep}{7pt}
    \renewcommand{\arraystretch}{0.9} 
    \begin{tabularx}{\linewidth}{lcccccc}
        \toprule
        \textbf{Model} & \textbf{\makecell[c]{Params\\(M)}} & \textbf{\makecell[c]{Test Acc.\\(\%)}} & \textbf{\makecell[c]{F1-Score\\(\%)}} & \textbf{\makecell[c]{Precision\\(\%)}} & \textbf{\makecell[c]{Sensitivity\\(\%)}} & \textbf{\makecell[c]{Kappa\\(\%)}} \\
        \midrule
        VGG19 \cite{simonyan2014very} & 143.6 & 19.38 & 17.14 & 19.26 & 19.38 & 7.86 \\
        EfficientNetB3 \cite{tan2019efficientnet} & 8.0 & 17.12 & 12.75 & 14.54 & 17.12 & 5.29 \\
        ResNet-50 \cite{he2016deep} & 14.3 & 16.38 & 11.55 & 10.72 & 16.38 & 4.43 \\
        XceptionNet \cite{xceptionnet} & 7.1 & 16.25 & 15.03 & 19.09 & 16.25 & 4.29 \\
        MobileNet V2 \cite{sandler2018mobilenetv2} & 55.8 & 16.12 & 10.79 & 13.70 & 16.12 & 4.14 \\
        VGG16 \cite{simonyan2014very} & 21.8 & 15.12 & 8.62 & 12.01 & 15.12 & 3.00 \\
        EfficientNetV2B0 \cite{tan2021efficientnetv2} & 60.2 & 15.12 & 10.30 & 14.33 & 15.12 & 3.00 \\
        DenseNet-201 \cite{huang2017densely} & 20.2 & 14.12 & 8.79 & 12.92 & 14.12 & 1.86 \\
        ViT-Base \cite{vit_main}& 22.9 & 13.38 & 11.33 & 17.96 & 13.38 & 1.00 \\
        ResNet101V2 \cite{he2016identity} & 44.6 & 13.25 & 10.64 & 12.52 & 13.25 & 0.86 \\
        ResNet-152V2 \cite{he2016identity} & 60.3 & 12.88 & 7.54 & 9.21 & 12.88 & 0.43 \\
        Inception-ResNetV2 \cite{szegedy2017inception} & 3.5 & 12.25 & 10.56 & 10.55 & 12.25 & -0.29 \\
        EfficientNetV2B3 \cite{tan2021efficientnetv2} & 14.5 & 12.25 & 6.87 & 7.75 & 12.25 & -0.29 \\
        ResNet-101 \cite{resnet50} & 44.5 & 12.12 & 10.26 & 10.68 & 12.12 & -0.43 \\
        EfficientNetB7 \cite{tan2019efficientnet} & 66.4 & 12.12 & 3.92 & 8.79 & 12.12 & -0.43 \\
        Inception V3 \cite{szegedy2016rethinking} & 23.8 & 11.88 & 8.61 & 10.82 & 11.88 & -0.71 \\
        ResNet-18 \cite{resnet50} & 11.7 & 11.75 & 9.47 & 20.93 & 11.75 & -0.86 \\
        DenseNet-121 \cite{huang2017densely} & 12.2 & 11.75 & 9.99 & 15.13 & 11.75 & -0.86 \\
        EfficientNetB0 \cite{tan2019efficientnet} & 5.3 & 11.12 & 9.09 & 10.70 & 11.12 & -1.57 \\
        ResNet-152 \cite{resnet50}  & 86.6 & 9.50 & 8.23 & 11.80 & 9.50 & -3.43 \\
        MobileNetV3Large \cite{howard2019searching} & 5.4 & 9.25 & 8.92 & 12.01 & 9.25 & -3.71 \\
        Inception V4 \cite{szegedy2017inception} & 42.7 & 8.88 & 6.99 & 15.35 & 8.88 & -4.14 \\
        DenseNet-169 \cite{huang2017densely} & 25.6 & 8.75 & 8.65 & 19.46 & 8.75 & -4.29 \\
        ResNet-34 \cite{resnet50}  & 138.4 & 8.12 & 6.49 & 10.20 & 8.12 & -5.00 \\
        ResNet50v2 \cite{he2016identity} & 25.6 & 8.12 & 5.25 & 5.93 & 8.12 & -5.00 \\
        MobileNetV3Small \cite{howard2019searching} & 2.5 & 7.38 & 4.82 & 3.74 & 7.38 & -5.86 \\
        \bottomrule
    \end{tabularx}
\end{table*}

    \begin{table*}[t]
        \caption{Performance of all DL architecture on stomach infection test dataset (5-Fold Evaluation) using Test Accuracy(\%), Test Loss, F1-Score(\%), Precision(\%) Sensitivity(\%), Specificity (\%), Kappa(\%)}
        \label{Performance of deep neural network models}
        \small
        \setlength{\tabcolsep}{4.1pt}
        \renewcommand{\arraystretch}{0.9} 
        \begin{tabularx}{\linewidth}{lcccccr}
            \toprule
            \textbf{Model} & \textbf{Test Accuracy (\%)} & \textbf{F1-Score (\%)} & \textbf{Precision (\%)} & \textbf{Sensitivity (\%)}  & \textbf{Specificity (\%)} & \textbf{Kappa (\%)}\\
            \midrule
            \textbf{CoATNet} & 83.77 $\pm$ 0.46 & 83.43 $\pm$ 0.51 & 83.78 $\pm$ 0.59 & 83.49 $\pm$ 0.52 & 82.30 $\pm$ 2.47 & 81.27 $\pm$ 0.60\\
            \midrule
            \textbf{EfficientNetB3} & 87.42 $\pm$ 0.21 & 87.07 $\pm$ 0.28 & 87.26 $\pm$ 0.31 & 87.07 $\pm$ 0.28 & 90.53 $\pm$ 2.33 & 85.37 $\pm$ 0.31\\
            \midrule
            \textbf{MobileNetV2} & 66.15 $\pm$ 1.00 & 66.09 $\pm$ 1.27 & 67.69 $\pm$ 0.98 & 66.16 $\pm$ 1.15 & 63.79 $\pm$ 7.14 & 61.47 $\pm$ 1.40\\
            \midrule
            \textbf{ResNet50} & 88.80 $\pm$ 1.71 & 88.37 $\pm$ 1.92 & 89.41 $\pm$ 1.19 & 88.59 $\pm$ 1.67 & 92.30 $\pm$ 5.26 & 87.11 $\pm$ 1.84\\
            \midrule
            \textbf{VGG16} & 80.53 $\pm$ 0.53 & 80.21 $\pm$ 0.62 & 80.56 $\pm$ 0.75 & 80.27 $\pm$ 0.65 & 81.75 $\pm$ 3.32 & 77.63 $\pm$ 0.74\\
            \midrule
            \textbf{VGG19} & 79.02 $\pm$ 1.11 & 78.60 $\pm$ 1.21 & 79.01 $\pm$ 1.28 & 78.63 $\pm$ 1.21 & 78.69 $\pm$ 3.40 & 75.76 $\pm$ 1.37\\
            \midrule
            \textbf{VIT} & 75.55 $\pm$ 2.00 & 74.92 $\pm$ 2.23 & 75.76 $\pm$ 1.57 & 75.18 $\pm$ 2.02 & 66.92 $\pm$ 11.66 & 71.89 $\pm$ 2.21\\
            \midrule
            \textbf{XceptionNet} & 92.04 $\pm$ 0.78 & 91.85 $\pm$ 0.82 & 92.09 $\pm$ 0.70 & 91.86 $\pm$ 0.81 & 92.78 $\pm$ 4.11 & 90.81 $\pm$ 0.91 \\
            \midrule
            \textbf{\makecell[l]{MobileCoAtNet\\(Proposed)}} & 97.53 $\pm$ 3.81 & 97.56 $\pm$ 3.81 & 97.62 $\pm$ 3.74 & 97.54 $\pm$ 3.82  & 97.52 $\pm$ 3.81 & 97.24 $\pm$ 4.31\\
            \bottomrule
        \end{tabularx}
        \end{table*}

    \begin{figure*}[!ht]
        \centering
        \includegraphics[width=0.5\linewidth]{
            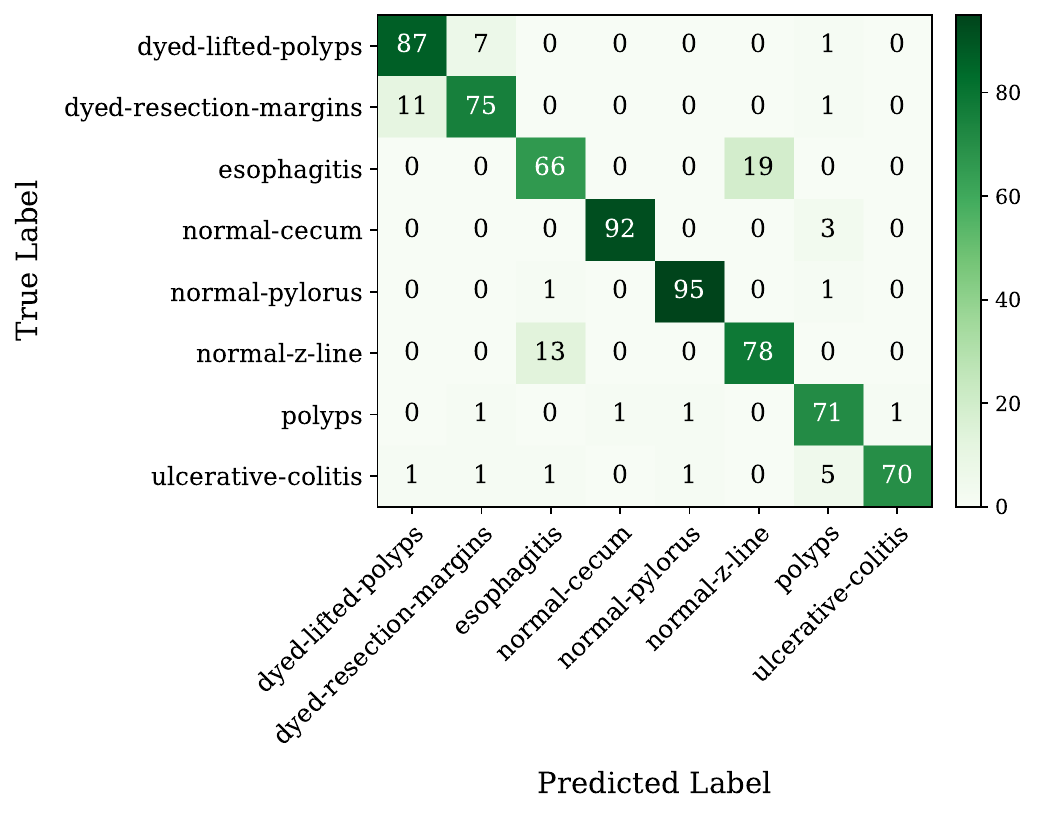
        }
        \caption{Confusion matrix for the proposed MobileCoAtNet model for the multi-class classification task.}
        \label{Proposed_Model_confusion_matrix}
    \end{figure*}

    \begin{figure*}[!ht]
        \centering
        \includegraphics[width=0.7\linewidth]{
            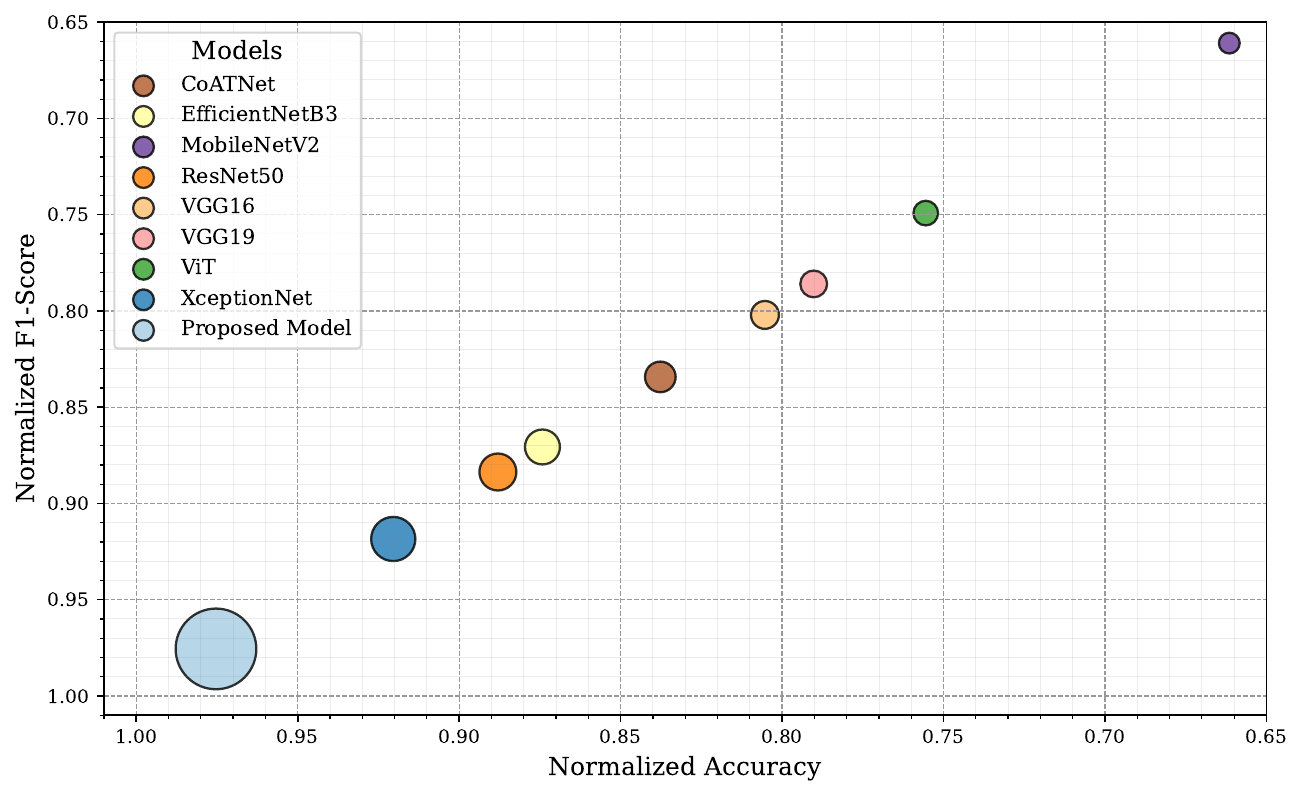
        }
        \caption{Accuracy vs. F1-score scatter plot for different model combinations for 128 batch size. Here, all of the models compete with each other to be the best-performing model. Best-performing model indicators are becoming larger in size as they are going towards the origin. The best-performing model is the one that is closest to the origin.}
        \label{Accuracy vs F1 Graph}
    \end{figure*}


\begin{table*}[!t]
\caption{Performance evaluation of the MobileCoAtNet model with 5-fold cross-validation for eight different classes of stomach disease.}
\label{table:Performance_MobileCoAtNet_5fold_cv}
\small
\centering
\setlength{\tabcolsep}{2.2pt} 
\renewcommand{\arraystretch}{1} 
\begin{tabularx}{\linewidth}{llccccccr}
        \toprule
            \multirow{2}{*}{\textbf{Network}} & \multirow{2}{*}{\textbf{Class}} & \multirow{2}{*}{\textbf{Fold}} & \textbf{Test Accuracy} & \textbf{F1} & \textbf{Sensitivity} & \textbf{Precision} & \textbf{Specificity} & \textbf{Kappa} \\
        & & & \textbf{(\%)} & \textbf{Score (\%)} & \textbf{(\%)} & \textbf{(\%)} & \textbf{(\%)} & \textbf{(\%)}\\
        \midrule
        & \multirow{5}{*}{\makecell[l]{8 (Dyed lifted polyps, Dyed\\resection margins, Esophagitis,\\Normal cecum, Normal pylorus,\\Normal z line, Polyps,\\Ulcerative colitis)}} & Fold 1 & 99.15 & 99.29 & 99.27 & 99.34 & 99.03 & 99.19 \\
        Proposed & & Fold 2 & 99.57 & 99.43 & 99.43 & 99.43 & 99.71 & 99.35 \\
        Model & & Fold 3 & 99.43 & 99.58 & 99.56 & 99.60 & 99.30 & 99.51 \\
        & & Fold 4 & 99.57 & 99.57 & 99.56 & 99.60 & 99.58 & 99.51 \\
        & & Fold 5 & 89.91 & 89.94 & 89.90 & 90.14 & 89.92 & 88.62 \\
        \hline
        & & Average & \makecell{97.53 \\ $\pm$ 3.81} & \makecell{97.56 \\ $\pm$ 3.81} & \makecell{97.54 \\ $\pm$ 3.82} & \makecell{97.62 \\ $\pm$ 3.74} & \makecell{97.52 \\ $\pm$ 3.81} & \makecell{97.24 \\ $\pm$ 4.31} \\
        \bottomrule
    \end{tabularx}
\end{table*}

\begin{table*}[!htb]
        \caption{Zero-shot-based initial LLM evaluation to select the best models for the clinical reasoning experiment.}
        \label{table:llm_selection}
        \centering
        \small
        \setlength{\tabcolsep}{0.67cm}
        \renewcommand{\arraystretch}{1}
        \begin{tabularx}{\linewidth}{lccc}
        \toprule
        \textbf{Model}&\textbf{BERT-Score}&\textbf{Cosine-Similarity}&\textbf{ROUGE-L}\\ 
        \midrule
        \makecell[l]{Deep Seek R1 (8B) \cite{deepseek_r1_2025}} & 65.28 & 13.57 & 8.09 \\ 
        \makecell[l]{Deep Seek R1 (14B) \cite{deepseek_r1_2025}} & 65.79 & 13.16 & 10.14 \\ 
        \makecell[l]{Deep Seek R1 (32B) \cite{deepseek_r1_2025}} & 65.51 & 13.99 & 10.81 \\ 
        \makecell[l]{Falcon (7B) \cite{falcon40b_2023}} & 65.64 & 12.11 & 10.11 \\ 
        \makecell[l]{Falcon-3 (7B) \cite{falcon40b_2023}} & 66.31 & 16.44 & 10.64 \\ 
        \makecell[l]{Mistral-Instruct (8B) \cite{jiang2023mistral7b}} & 66.40 & 14.99 & 10.67 \\ 
        \makecell[l]{Mistral-Instruct (7B) \cite{jiang2023mistral7b}} & 66.01 & 13.72 & 10.78 \\ 
        \makecell[l]{Llama-3.1-Instruct (8B) \cite{llama32024}} & 65.70 & 15.51 & 9.41 \\ 
        \makecell[l]{Llama-3.2-Instruct (1B) \cite{llama32024}} & 64.05 & 10.83 & 6.94 \\ 
        \makecell[l]{Llama-3.2-Instruct (3B) \cite{llama32024}} & 64.90 & 12.44 & 7.97 \\ 
        \makecell[l]{Llama-3.2-Instruct (11B) \cite{llama32024}} & 66.00 & 17.06 & 10.01 \\ 
        \makecell[l]{Qwen-2.5-Instruct (1.5B) \cite{qwen2_2024}} & 65.28 & 15.58 & 7.99 \\ 
        \makecell[l]{Qwen-2.5-Instruct (7B) \cite{qwen2_2024}} & 66.41 & 15.66 & 11.02 \\ 
        \makecell[l]{Qwen-3 (4B) \cite{qwen3_2025}} & 64.55 & 12.29 & 7.95 \\ 
        \makecell[l]{Qwen-3 (8B) \cite{qwen3_2025}} & 65.16 & 15.02 & 9.03 \\ 
        \makecell[l]{Qwen-3 (14B) \cite{qwen3_2025}} & 65.67 & 17.13 & 9.04 \\ 
        \makecell[l]{Gemma-2 (2B) \cite{gemma2_2024}} & 62.53 & 6.33 & 4.48 \\ 
        \makecell[l]{Gemma-2 (9B) \cite{gemma2_2024}} & 64.68 & 13.59 & 8.35 \\ 
        \makecell[l]{Gemini-1.5-Flash (8B) \cite{gemini15_2023}} & 65.89 & 14.84 & 11.66 \\ 
        \makecell[l]{Gemini-1.5-Flash \cite{gemini15_2023}} & 66.24 & 16.59 & 11.83 \\ 
        \makecell[l]{Gemini-1.5-Pro \cite{gemini15_2023}} & 65.69 & 17.39 & 11.19 \\ 
        \makecell[l]{Gemini-2.0-Flash-Lite \cite{gemini20_2024}} & 65.38 & 14.32 & 10.77 \\ 
        \makecell[l]{Gemini-2.0-Flash \cite{gemini20_2024}} & 65.56 & 17.13 & 10.45 \\ 
        \makecell[l]{Gemini-2.5-Flash \cite{gemini25_2025}} & 65.38 & 16.66 & 9.26 \\ 
        \makecell[l]{Gemini-2.5-Pro \cite{gemini25_2025}} & 64.94 & 17.09 & 9.00 \\ 
        \makecell[l]{O3 \cite{brown2020language}} & 70.20 & 35.02 & 31.37 \\ 
        \makecell[l]{O3-Mini \cite{brown2020language}} & 66.42 & 21.27 & 13.91 \\ 
        \makecell[l]{O4-Mini \cite{brown2020language}} & 68.95 & 32.70 & 29.83 \\ 
        \makecell[l]{GPT-3.5-Turbo \cite{gpt4_technical_report_2023}} & 71.59 & 51.70 & 51.55 \\ 
        \makecell[l]{GPT-4.1-Nano \cite{gpt4_technical_report_2023}} & 71.02 & 44.35 & 44.54 \\ 
        \makecell[l]{GPT-4.1-Mini \cite{gpt4_technical_report_2023}} & 70.48 & 40.19 & 36.29 \\ 
        \makecell[l]{GPT-4.1 \cite{gpt4_technical_report_2023}} & 72.26 & 45.61 & 44.41 \\
        \bottomrule
        \end{tabularx}
    \end{table*}
\begin{table*}[!ht]
\centering
\small
\caption{Ablation study on CRS$_{\text{prompt}}$ scores across different weight combinations to select the best combination.}
\label{tab:crs_prompt_weights}
\setlength{\tabcolsep}{1.6pt}
\renewcommand{\arraystretch}{1}

\begin{tabular}{lcccccccc}
\toprule
\textbf{\makecell[l]{Weight setting\\($w_1$,$w_2$,$w_3$)}} 
& \textbf{Gemini-1.5} 
& \textbf{Gemini-1.5-8B}
& \textbf{Gemini-1.5-Pro}
& \textbf{Gemini-2.0}
& \textbf{GPT-3.5}
& \textbf{GPT-4.1}
& \textbf{GPT-4.1-mini}
& \textbf{GPT-4.1-nano}
\\
\midrule

$(0.2,0.3,0.5)$ & 0.1864 & 0.1990 & 0.1724 & 0.1960 & 0.1957 & 0.2436 & 0.2476 & 0.1946 \\
$(0.2,0.4,0.4)$ & 0.1754 & 0.1874 & 0.1624 & 0.1844 & 0.1837 & 0.2287 & 0.2331 & 0.1835 \\
$(0.2,0.5,0.3)$ & 0.1645 & 0.1758 & 0.1524 & 0.1727 & 0.1717 & 0.2138 & 0.2187 & 0.1723 \\

$(0.3,0.2,0.5)$ & 0.2104 & 0.2242 & 0.1954 & 0.2218 & 0.2176 & 0.2758 & 0.2778 & 0.2191 \\
$(0.3,0.3,0.4)$ & 0.1995 & 0.2126 & 0.1854 & 0.2101 & 0.2056 & 0.2609 & 0.2634 & 0.2080 \\
$(0.3,0.4,0.3)$ & 0.1885 & 0.2010 & 0.1754 & 0.1985 & 0.1936 & 0.2460 & 0.2489 & 0.1969 \\
$(0.3,0.5,0.2)$ & 0.1776 & 0.1894 & 0.1654 & 0.1869 & 0.1816 & 0.2311 & 0.2344 & 0.1858 \\

$(0.4,0.2,0.4)$ & 0.2235 & 0.2378 & 0.2084 & 0.2359 & 0.2275 & 0.2930 & 0.2936 & 0.2326 \\
$(0.4,0.3,0.3)$ & 0.2126 & 0.2262 & 0.1984 & 0.2243 & 0.2155 & 0.2782 & 0.2791 & 0.2214 \\
$(0.4,0.4,0.2)$ & 0.2016 & 0.2147 & 0.1884 & 0.2127 & 0.2036 & 0.2633 & 0.2647 & 0.2103 \\

$\boldsymbol{(0.5,0.2,0.3)}$ & \textbf{0.2366} & \textbf{0.2514} & \textbf{0.2214} & \textbf{0.2501} & \textbf{0.2374} & \textbf{0.3103} & \textbf{0.3094} & \textbf{0.2460} \\
$(0.5,0.3,0.2)$ & 0.2257 & 0.2399 & 0.2114 & 0.2384 & 0.2255 & 0.2954 & 0.2949 & 0.2349 \\

\bottomrule
\end{tabular}
\end{table*}

\section{Results and Findings}
\label{Experimental Results}

\begin{figure*}[t]
    \centering
    \includegraphics[width=0.9\linewidth]{
        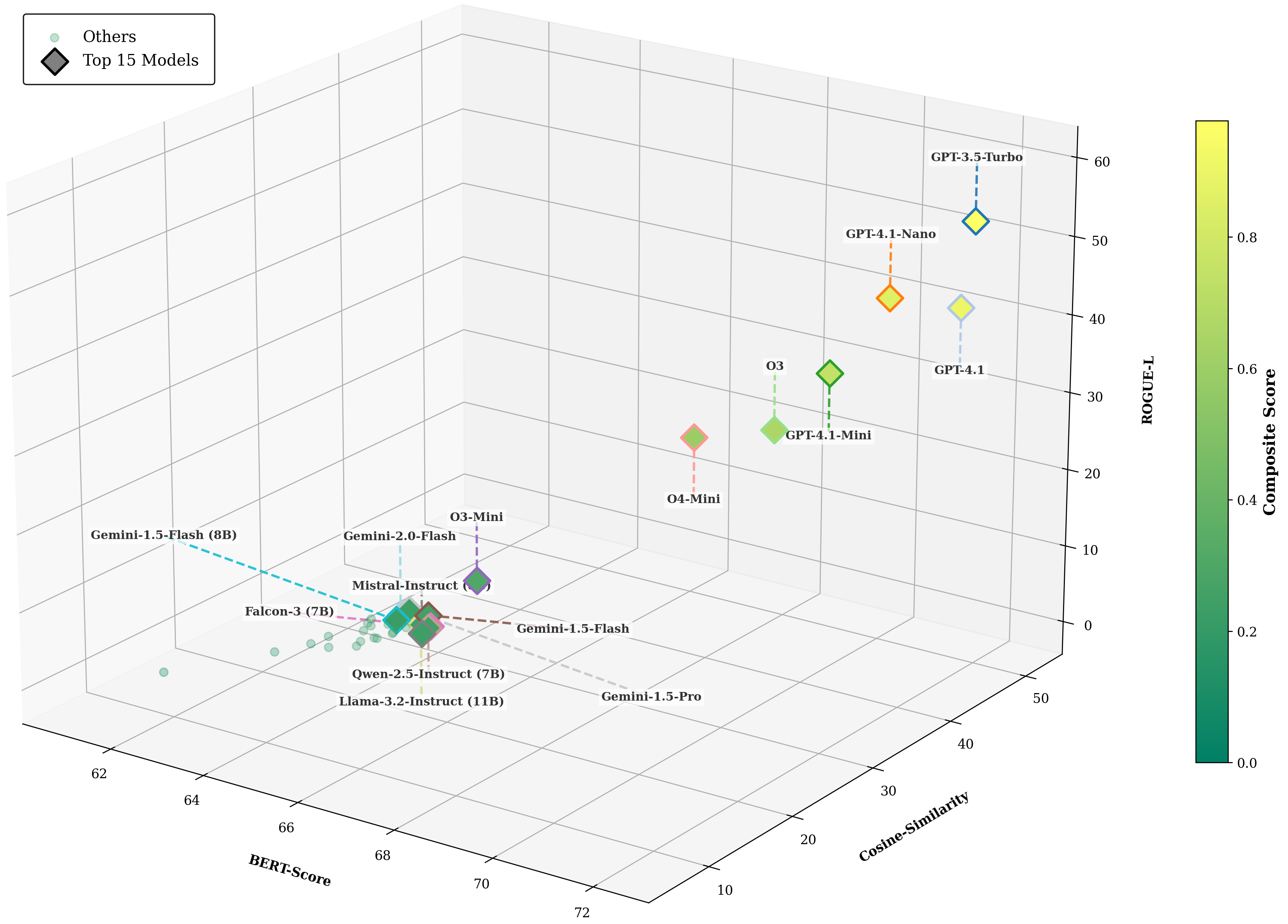
    }
    \caption{Top 15 LLM selection using the composite score, which is calculated using four CRS$_{\text{prompt}}$ scores.}
    \label{fig:llm_selection_composite_score_graph}
\end{figure*} 

    \begin{table*}[t]
        \centering
        \small
        \caption{Final leaderboard for DL\textsuperscript{3}M-based clinical reasoning generation.}
        \label{table:llm_reasoning_leaderboard_reversed}
        \setlength{\tabcolsep}{8pt}
        \renewcommand{\arraystretch}{1.1} 
        \begin{tabularx}{\textwidth}{c l c c c c c r}
        \toprule
        \textbf{Rank} & \textbf{Model} & \textbf{CRS (P1)} & \textbf{CRS (P2)} & \textbf{CRS (P3)} & \textbf{CRS (P4)} & \textbf{Final Score} & \textbf{Category} \\
        \midrule
        1 & \textbf{GPT 4.1} & 0.537 & 0.529 & 0.519 & 0.523 & \textbf{0.530} & \textbf{Moderate-CRA} \\
        2 & \textbf{GPT 4.1 Mini} & 0.531 & 0.527 & 0.485 & 0.481 & \textbf{0.515} & \textbf{Moderate-CRA} \\
        3 & \textbf{Gemini 1.5 Flash 8b} & 0.462 & 0.474 & 0.477 & 0.487 & \textbf{0.471} & \textbf{Low-CRA} \\
        4 & \textbf{GPT 3.5 turbo} & 0.464 & 0.459 & 0.483 & 0.496 & \textbf{0.470} & \textbf{Low-CRA} \\
        5 & \textbf{GPT 4.1 Nano} & 0.440 & 0.469 & 0.503 & 0.498 & \textbf{0.467} & \textbf{Low-CRA} \\
        6 & \textbf{Gemini 2.0 Flash} & 0.454 & 0.472 & 0.482 & 0.454 & \textbf{0.465} & \textbf{Low-CRA} \\
        7 & \textbf{O3} & 0.462 & 0.462 & 0.459 & 0.471 & \textbf{0.462} & \textbf{Low-CRA} \\
        8 & \textbf{Gemini 1.5 Pro} & 0.445 & 0.447 & 0.497 & 0.480 & \textbf{0.460} & \textbf{Low-CRA} \\
        9 & \textbf{Gemini 1.5 Flash} & 0.426 & 0.460 & 0.473 & 0.482 & \textbf{0.451} & \textbf{Low-CRA} \\
        10 & \textbf{O4-Mini} & 0.445 & 0.449 & 0.443 & 0.276 & \textbf{0.429} & \textbf{Low-CRA} \\
        11 & \textbf{O3-Mini} & 0.234 & 0.234 & 0.234 & 0.234 & \textbf{0.234} & \textbf{Low-CRA} \\
        \bottomrule
        \end{tabularx}
    \end{table*}

In our experiment, we used two types of models to complete our framework- one inspired from DL and the other by LLM. Initialy we select a list of models and found the best performing one after evaluating on task-specific benchmark data sets. 

\subsection{Performance Evaluation of DL Models}
\label{Analysis of the best performing DL}
\textbf{A. DL Model Selection}\\

    At first, we selected twenty-five DL models from nine different architecture families based on parameter size. We selected a subset of 500 data samples from our dataset \cite{kvasir}. We used a set of evaluation metrics: test accuracy, F1 score, precision, sensitivity, and Cohen's kappa. We evaluated the models on the selected data sample using a zero-shot mechanism. We sorted the models based on F1-score and selected the best-performing eight models for further experimentation.

\begin{figure*}[!ht]
    \centering
    \begin{subfigure}[!t]{0.75\linewidth}
    \centering
    \includegraphics[width=\linewidth]{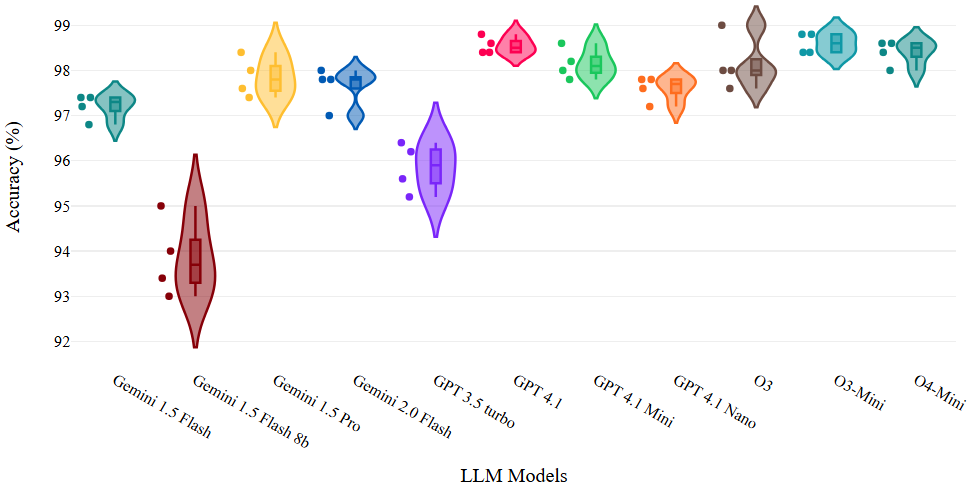}
    \caption{Prompt sensitivity in LLM for four different directive levels for the multiple-choice task.}
    \label{fig:classification_llm_violinplot_accuracy}
\end{subfigure}
    
\begin{subfigure}[!ht]{0.75\linewidth}
    \centering
    \includegraphics[width=\linewidth]{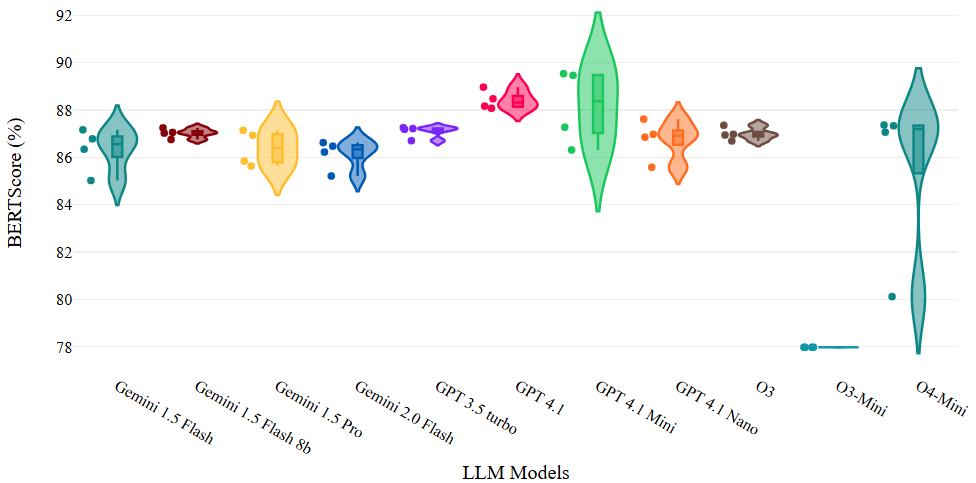}
    \caption{Prompt sensitivity in LLM for four different directive levels for the multi-step reasoning task.}
    \label{fig:reasoning_llm_violinplot_BERTScore}
\end{subfigure}
    
        \caption{Violin plot for prompt sensitivity evaluation in (a) multiple-choice and (b) reasoning tasks.}
        \label{llm_violinplot}
    \end{figure*}

    
\begin{sidewaystable*}[!ht]
\centering
\small
\caption{Multiple-choice performance of fifteen LLMs across four levels of prompts}
\label{table:llm_performance_classification}
\begin{tabular}{llrrrrrrrrrrrr}
    \toprule
    \multirow{2}{*}{Model} & \multirow{2}{*}{Class} &
    \multicolumn{3}{c}{Prompt 1} &
    \multicolumn{3}{c}{Prompt 2} &
    \multicolumn{3}{c}{Prompt 3} &
    \multicolumn{3}{c}{Prompt 4} \\
    & & 
    Acc & F1 & Kappa &
    Acc & F1 & Kappa &
    Acc & F1 & Kappa &
    Acc & F1 & Kappa \\
    \midrule
    \multirow{5}{*}{GPT 4.1} & Polyps &  0.9600 & 0.8916 &    0.9219 &  0.9500 & 0.8527 &    0.9028 &  0.9700 & 0.9545 &    0.9415 &  0.9700 & 0.9545 & 0.9415 \\
    & Ulcerative Colitis &  0.9800 & 0.7929 & 0.9672 &  0.9800 & 0.9000 &    0.9671 &  0.9600 & 0.4869 &    0.9346 &  0.9500 & 0.4158 &    0.9189 \\
    & Esophagitis &  1.0000 & 1.0000 & 1.0000 &  1.0000 & 1.0000 & 1.0000 &  0.9900 & 0.7989 &    0.9483 &  1.0000 & 1.0000 & 1.0000 \\
    & Healthy Class &  1.0000 & 1.0000 & 1.0000 &  1.0000 & 1.0000 &    1.0000 &  1.0000 & 1.0000 &    1.0000 &  1.0000 & 1.0000 & 1.0000 \\
    & Removal of Lesions &  1.0000 & 1.0000 &    1.0000 &  1.0000 & 1.0000 &    1.0000 &  1.0000 & 1.0000 &    1.0000 &  1.0000 & 1.0000 & 1.0000 \\
    \hline
    
    \multirow{5}{*}{GPT 3.5 turbo} & Polyps &  0.8600 & 0.6976 & 0.7126 &  0.8900 & 0.7833 &    0.7814 &  0.8800 & 0.7518 &    0.7490 &  0.8800 & 0.5972 &    0.7551 \\
    & Ulcerative Colitis &  0.9500 & 0.9497 &    0.9180 &  0.9400 & 0.9383 &    0.9022 &  0.9400 & 0.6912 &    0.9014 &  0.9300 & 0.7002 &    0.8846 \\
    &        Esophagitis &  1.0000 & 1.0000 &    1.0000 &  0.9900 & 0.8974 &    0.9460 &  0.9800 & 0.9556 &    0.8810 &  0.9700 & 0.7056 &    0.8214 \\
    &      Healthy Class &  1.0000 & 1.0000 &    1.0000 &  1.0000 & 1.0000 &    1.0000 &  1.0000 & 1.0000 &    1.0000 &  1.0000 & 1.0000 &    1.0000 \\
    &      Removal of Lesions &  1.0000 & 1.0000 &    1.0000 &  1.0000 & 1.0000 &    1.0000 &  0.9600 & 0.4898 &    1.0000 &  1.0000 & 1.0000 &    1.0000 \\
    \hline
     
    \multirow{5}{*}{GPT 4.1 Mini} & Polyps &  0.9600 & 0.9058 &    0.9212 &  0.9500 & 0.8858 &    0.9018 &  0.9500 & 0.8648 &    0.9013 &  0.9600 & 0.8828 &    0.9208 \\
    & Ulcerative Colitis &  0.9800 & 0.9000 &    0.9671 &  0.9800 & 0.9000 &    0.9671 &  0.9800 & 0.9000 &    0.9671 &  0.9800 & 0.9000 &    0.9671 \\
    &        Esophagitis &  0.9900 & 0.8974 &    0.9460 &  0.9900 & 0.8974 &    0.9460 &  0.9800 & 0.8736 &    0.8869 &  0.9900 & 0.9794 &    0.9433 \\
    & Healthy Class &  0.9600 & 0.5486 &    0.4872 &  0.9800 & 0.8282 &    0.6575 &  1.0000 & 1.0000 &    1.0000 &  1.0000 & 1.0000 &    1.0000 \\
    & Removal of Lesions &  1.0000 & 1.0000 &    1.0000 &  1.0000 & 1.0000 &    1.0000 &  1.0000 & 1.0000 &    1.0000 &  1.0000 & 1.0000 &    1.0000 \\
    \hline
      
    \multirow{5}{*}{GPT 4.1 Nano} & Polyps &  0.9400 & 0.8850 &    0.8812 &  0.9500 & 0.9265 &    0.9006 &  0.9400 & 0.8708 &    0.8832 &  0.9600 & 0.9397 &    0.9228 \\
    & Ulcerative Colitis &  0.9600 & 0.8498 &    0.9347 &  0.9500 & 0.8440 &    0.9183 &  0.9500 & 0.8217 &    0.9189 &  0.9600 & 0.8425 &    0.9349 \\
    &        Esophagitis &  1.0000 & 1.0000 &    1.0000 &  0.9900 & 0.9794 &    0.9433 &  0.9900 & 0.9794 &    0.9433 &  0.9900 & 0.9794 &    0.9433 \\
    & Healthy Class &  0.9800 & 0.8282 &    0.6575 &  1.0000 & 1.0000 & 1.0000 &  0.9800 & 0.8282 &    0.6575 &  0.9800 & 0.8282 & 0.6575 \\
    & Removal of Lesions &  1.0000 & 1.0000 &    1.0000 &  1.0000 & 1.0000 &    1.0000 &  1.0000 & 1.0000 &    1.0000 &  1.0000 & 1.0000 &    1.0000 \\
    \hline
      
    \multirow{5}{*}{O3} & Polyps & 0.9400 & 0.8330 & 0.8839 & 0.9500 & 0.8399 & 0.9025 & 0.9400 & 0.8142 & 0.8826 &  0.9700 & 0.9162 & 0.9409 \\
    & Ulcerative Colitis & 0.9700 & 0.7234 & 0.9505 & 0.9600 & 0.7177 & 0.9341 & 0.9700 & 0.8943 &    0.9506 & 0.9900 & 0.9942 & 0.9835 \\
    & Esophagitis &  0.9900 & 0.8974 &    0.9460 &  0.9900 & 0.8974 &    0.9460 &  0.9700 & 0.8222 &    0.8384 &  0.9900 & 0.8974 &    0.9460 \\
    & Healthy Class &  1.0000 & 1.0000 &    1.0000 &  1.0000 & 1.0000 &    1.0000 &  1.0000 & 1.0000 &    1.0000 &  1.0000 & 1.0000 &    1.0000 \\
    & Removal of Lesions &  1.0000 & 1.0000 &    1.0000 &  1.0000 & 1.0000 &    1.0000 &  1.0000 & 1.0000 &    1.0000 &  1.0000 & 1.0000 &    1.0000 \\
    \hline

    \multirow{5}{*}{O4-Mini} & Polyps &  0.9700 & 0.8616 &    0.9413 &  0.9600 & 0.8543 &    0.9211 &  0.9700 & 0.9068 &    0.9417 &  0.9700 & 0.9471 &    0.9406 \\
    & Ulcerative Colitis &  0.9800 & 0.7342 &    0.9670 &  0.9700 & 0.8943 &    0.9506 &  0.9800 & 0.9000 &    0.9671 &  0.9800 & 0.9807 &    0.9668 \\
    & Esophagitis &  0.9800 & 0.8460 &    0.8969 &  0.9700 & 0.8222 &    0.8384 &  0.9700 & 0.8562 &    0.8380 &  0.9800 & 0.8736 &    0.8869 \\
    & Healthy Class &  1.0000 & 1.0000 &    1.0000 &  1.0000 & 1.0000 &    1.0000 &  1.0000 & 1.0000 &    1.0000 &  1.0000 & 1.0000 &    1.0000 \\
    & Removal of Lesions &  1.0000 & 1.0000 &    1.0000 &  1.0000 & 1.0000 &    1.0000 &  1.0000 & 1.0000 &    1.0000 &  1.0000 & 1.0000 &    1.0000 \\
    \hline
    
    \multirow{5}{*}{O3-Mini} &             Polyps &  0.9700 & 0.8988 &    0.9419 &  0.9700 & 0.8988 &    0.9419 &  0.9700 & 0.8988 &    0.9419 &  0.9600 & 0.8916 &    0.9219 \\
    & Ulcerative Colitis &  0.9800 & 0.9000 &    0.9671 &  0.9800 & 0.9000 &    0.9671 &  0.9700 & 0.8556 &    0.9510 &  0.9800 & 0.9804 &    0.9668 \\
    &        Esophagitis &  0.9900 & 0.8974 &    0.9460 &  0.9900 & 0.8974 &    0.9460 &  0.9800 & 0.8736 &    0.8869 &  0.9800 & 0.8736 &    0.8869 \\
    & Healthy Class &  1.0000 & 1.0000 &    1.0000 &  1.0000 & 1.0000 &    1.0000 &  1.0000 & 1.0000 &    1.0000 &  1.0000 & 1.0000 &    1.0000 \\
    &      Removal of Lesions &  1.0000 & 1.0000 &    1.0000 &  1.0000 & 1.0000 &    1.0000 &  1.0000 & 1.0000 &    1.0000 &  1.0000 & 1.0000 &    1.0000 \\
    \hline

    \multirow{5}{*}{Gemini 1.5 Flash} & Polyps &  0.9100 & 0.7899 & 0.8239 & 0.8900 & 0.7527 & 0.7847 &  0.9100 & 0.7899 & 0.8239 &  0.9200 & 0.8495 & 0.8436 \\
    & Ulcerative Colitis & 0.9600 & 0.8498 & 0.9346 & 0.9600 & 0.8811 & 0.9344 &  0.9700 & 0.8940 & 0.9506 &  0.9700 & 0.8943 & 0.9506 \\
    & Esophagitis &  0.9900 & 0.9794 & 0.9433 &  0.9900 & 0.9794 & 0.9433 &  0.9900 & 0.8974 &    0.9460 &  0.9800 & 0.8736 &    0.8869 \\
    & Healthy Class &  1.0000 & 1.0000 & 1.0000 &  1.0000 & 1.0000 & 1.0000 &  1.0000 & 1.0000 &    1.0000 &  1.0000 & 1.0000 &    1.0000 \\
    & Removal of Lesions &  1.0000 & 1.0000 &    1.0000 &  1.0000 & 1.0000 & 1.0000 &  1.0000 & 1.0000 & 1.0000 &  1.0000 & 1.0000 & 1.0000 \\
\bottomrule
\end{tabular}
\end{sidewaystable*}

\begin{sidewaystable*}[!ht]
\centering
\small
\caption{Multiple-choice performance of fifteen LLMs across four levels of prompts (Continued)}
\label{table:llm_performance_classification_2}
\begin{tabular}{llrrrrrrrrrrrr}
    \toprule
    \multirow{2}{*}{Model} & \multirow{2}{*}{Class} &
    \multicolumn{3}{c}{Prompt 1} &
    \multicolumn{3}{c}{Prompt 2} &
    \multicolumn{3}{c}{Prompt 3} &
    \multicolumn{3}{c}{Prompt 4} \\
    & & 
    Acc & F1 & Kappa &
    Acc & F1 & Kappa &
    Acc & F1 & Kappa &
    Acc & F1 & Kappa \\
    \midrule
    \multirow{5}{*}{Gemini 1.5 Pro} & Polyps &  0.9200 & 0.8021 &    0.8415 &  0.9400 & 0.8295 &    0.8829 &  0.9200 & 0.7910 &    0.8412 &  0.9100 & 0.7681 &    0.8219 \\
    & Ulcerative Colitis &  0.9800 & 0.9804 & 0.9668 &  0.9900 & 0.9942 &    0.9835 &  0.9700 & 0.8943 &    0.9506 &  0.9700 & 0.9677 &    0.9505 \\
    & Esophagitis &  1.0000 & 1.0000 & 1.0000 &  0.9900 & 0.9794 &    0.9433 &  0.9900 & 0.9794 &    0.9433 &  0.9900 & 0.9794 &    0.9433 \\
    & Healthy Class &  1.0000 & 1.0000 & 1.0000 &  1.0000 & 1.0000 &    1.0000 &  1.0000 & 1.0000 &    1.0000 &  1.0000 & 1.0000 &    1.0000 \\
    & Removal of Lesions &  1.0000 & 1.0000 & 1.0000 &  1.0000 & 1.0000 &    1.0000 &  1.0000 & 1.0000 &    1.0000 &  1.0000 & 1.0000 &    1.0000 \\
    \hline

    \multirow{5}{*}{Gemini 1.5 Flash 8b} & Polyps &  0.8100 & 0.6430 &    0.5716 &  0.8000 & 0.5914 &    0.5432 &  0.7900 & 0.5162 &    0.5021 &  0.7900 & 0.5177 &    0.5088 \\
    & Ulcerative Colitis &  0.9600 & 0.7221 &    0.9333 &  0.9200 & 0.8468 &    0.8671 &  0.9200 & 0.9036 &    0.8649 &  0.8900 & 0.8724 &    0.8133 \\
    & Esophagitis &  0.9800 & 0.9556 &    0.8810 &  0.9800 & 0.9615 &    0.8917 &  0.9600 & 0.8431 &    0.7349 &  0.9700 & 0.8709 &    0.8116 \\
    & Healthy Class &  1.0000 & 1.0000 &    1.0000 &  1.0000 & 1.0000 &    1.0000 &  1.0000 & 1.0000 & 1.0000 &  1.0000 & 1.0000 &    1.0000 \\
    & Removal of Lesions &  1.0000 & 1.0000 &    1.0000 &  1.0000 & 1.0000 &    1.0000 &  1.0000 & 1.0000 & 1.0000 &  1.0000 & 1.0000 &    1.0000 \\
    \hline

    \multirow{5}{*}{Gemini 2.0 Flash} & Polyps &  0.9300 & 0.8199 & 0.8592 &  0.9300 & 0.8678 & 0.8602 &  0.9400 & 0.8746 & 1.0000 & 0.9500 & 0.8842 & 0.9016 \\
    & Ulcerative Colitis &  0.9700 & 0.8556 & 0.9510 &  0.9700 & 0.7234 & 0.9505 &  0.9700 & 0.8873 &    0.9509 & 0.9400 & 0.6573 & 0.9040 \\
    & Esophagitis & 0.9900 & 0.9794 & 0.9433 & 0.9900 & 0.9794 & 0.9433 &  0.9900 & 0.9794 & 0.9433 &  0.9800 & 0.7824 & 0.8921 \\
    & Healthy Class & 1.0000 & 1.0000 & 1.0000 & 1.0000 & 1.0000 & 1.0000 & 1.0000 & 1.0000 & 1.0000 & 0.9800 & 0.8282 & 0.6575 \\
    &  Removal of Lesions &  1.0000 & 1.0000 &    1.0000 &  1.0000 & 1.0000 &  1.0000 &  1.0000 & 1.0000 &    1.0000 &  1.0000 & 1.0000 & 1.0000 \\
    \hline
    
    \multirow{5}{*}{Tiiuae/Falcon3 7b Base} & Polyps &  0.5100 & 0.3691 & 0.3037 & 0.6700 & 0.4438 & 0.4697 & 0.3900 & 0.2366 & 0.1782 & 0.3600 & 0.1843 & 0.1160\\
    & Ulcerative Colitis &  0.5800 & 0.4178 & 0.4239 & 0.8000 & 0.5542 & 0.6960 & 0.3300 & 0.3271 & 0.2026 & 0.2700 & 0.1810 & 0.0982 \\
    &        Esophagitis &  0.4700 & 0.2096 & 0.0908 & 0.8000 & 0.4807 & 0.4331 & 0.3800 & 0.2097 & 0.0975 & 0.3100 & 0.1737 & 0.0483 \\
    & Healthy Class &  0.6400 & 0.1580 & 0.0333 & 0.7200 & 0.1694 & 0.0463 & 0.4600 & 0.2799 & 0.0486 & 0.2600 & 0.0839 & 0.0043 \\
    &      Removal of Lesions &  0.4400 & 0.1528 & 0.0000 & 0.6400 & 0.1561 & 0.0000 & 0.3600 & 0.1324 & 0.0000 & 0.4800 & 0.1622 & 0.0000 \\
    \hline
    
    \multirow{5}{*}{Ministral 8B} & Polyps &  0.1900 & 0.2550 & 0.0810 & 0.5300 & 0.5276 & 0.3369 & 0.5100 & 0.3411 & 0.2326 & 0.6600 & 0.3860 & 0.3665 \\
    & Ulcerative Colitis &  0.4200 & 0.5299 & 0.2926 & 0.4900 & 0.5675 & 0.3526 & 0.5400 & 0.3719 & 0.3711 & 0.7400 & 0.5593 & 0.6084 \\
    &        Esophagitis &  0.3500 & 0.3101 & 0.0714 & 0.5000 & 0.3476 & 0.0871 & 0.6000 & 0.2264 & 0.1063 & 0.8400 & 0.6167 & 0.4554 \\
    & Healthy Class &  0.2400 & 0.1311 & 0.0058 & 0.4000 & 0.5196 & 0.0434 & 0.5800 & 0.5758 & 0.0683 & 0.7400 & 0.3784 & 0.1144 \\
    &      Removal of Lesions &  0.2800 & 0.2188 & 0.0000 & 0.2000 & 0.1111 & 0.0000 & 0.4400 & 0.2037 & 0.0000 & 0.6000 & 0.3750 & 0.0000 \\
    \hline
    
    \multirow{5}{*}{Meta Lama 3.2} & Polyps &  0.3100 & 0.1988 & 0.0556 & 0.6000 & 0.3947 & 0.3398 & 0.6200 & 0.3527 & 0.3513 & 0.6300 & 0.3104 & 0.3534 \\
    & Ulcerative Colitis &  0.3200 & 0.2134 & 0.0969 & 0.6100 & 0.4468 & 0.4404 & 0.6900 & 0.5060 & 0.5502 & 0.5500 & 0.3607 & 0.3548 \\
    & Esophagitis &  0.3600 & 0.1859 & 0.0698 & 0.5600 & 0.2422 & 0.1240 & 0.7700 & 0.3886 & 0.3328 & 0.6600 & 0.2760 & 0.1764 \\
    & Healthy Class &  0.3400 & 0.1030 & 0.0030 & 0.4200 & 0.1604 & 0.0410 & 0.7200 & 0.2917 & 0.1026 & 0.7000 & 0.1647 & -0.0218 \\
    & Removal of Lesions &  0.4800 & 0.1622 & 0.0000 & 0.7200 & 0.1674 & 0.0000 & 0.5600 & 0.1436 & 0.0000 & 0.6400 & 0.1561 & 0.0000 \\
    \hline
    
    \multirow{5}{*}{Qwen3-14B} & Polyps &  0.2000 & 0.1578 & 0.0741 & 0.2900 & 0.2500 & 0.1419 & 0.2200 & 0.1714 & 0.1079 & 0.2900 & 0.2656 & 0.1539 \\
    & Ulcerative Colitis &  0.3400 & 0.2836 & 0.2272 & 0.3400 & 0.3498 & 0.2384 & 0.1500 & 0.1374 & 0.0818 & 0.2400 & 0.2174 & 0.1566 \\
    & Esophagitis &  0.4800 & 0.2933 & 0.1224 & 0.4400 & 0.4253 & 0.1223 & 0.2600 & 0.2417 & 0.0729 & 0.3000 & 0.3145 & 0.0727 \\
    & Healthy Class &  0.4400 & 0.1549 & 0.0155 & 0.4400 & 0.4000 & 0.0476 & 0.2000 & 0.0678 & 0.0050 & 0.2000 & 0.0678 & 0.0045 \\
    & Removal of Lesions &  0.6400 & 0.1951 & 0.0000 & 0.5600 & 0.1795 & 0.0000 & 0.3600 & 0.1765 & 0.0000 & 0.5200 & 0.2281 & 0.0000 \\
\bottomrule
\end{tabular}
\end{sidewaystable*}

\textbf{B. Best Performing DL Model Selection}

We performed five-fold cross-validation on the selected eight models from zero-shot based model selection along with our proposed MobileCoAtNet on stomach infection test dataset (\ref{Performance of deep neural network models}).
    
The proposed MobileCoAtNet model along with eight selected models, was fine-tuned and evaluated on eight classes: dyed lifted polyps, dyed resection margins, esophagitis, normal cecum, normal pylorus, normal Z-line, Polyps, and ulcerative colitis. As summarised in \autoref{table:Performance_MobileCoAtNet_5fold_cv}, the model demonstrates consistently high performance across the first four folds, achieving test accuracies between 99.15\% and 99.57\%, with corresponding F1 scores ranging from 99.29\% to 99.58\% and low test losses between 0.0209 and 0.0475. The confusion matrix in \autoref{Proposed_Model_confusion_matrix} shows a strong diagonal dominance, indicates most samples are classified correctly and the proposed model can separate the gastrointestinal classes well.  In contrast, performance in Fold five is comparatively lower, with a test accuracy of 89.91\%, and an F1 score of 89.94\%. Our Proposed model achieved the highest score with an average test accuracy of \(97.53 \pm 3.81\%\) and mean F1 score \(97.56 \pm 3.81\%\) across five different folds.

After evaluating other 8 models XceptionNet was found as second best performer with test accuracy of (92.04\% $\pm$ 0.78) and F1 score of (91.85\% $\pm$ 0.82). The performance comparison of nine selected models including our proposed one (MobileCoAtNet) shows that MobileCoAtNet achieved the strongest performance with a large performance gap even with the second best performing model, which is evident by \autoref{Accuracy vs F1 Graph} (normalized Accuracy vs normalized F1 Score).

\label{Quantitative Analysis}%
The proposed MobileCoAtNet achieves the highest accuracy at 97.53\% and operates at 50.10 GigaFLOPs with 347.80M parameters, which is expected for a model optimized for maximum performance. Among the efficient models, MobileNetV2 and EfficientNetB3 stand out. MobileNetV2 has the lowest computational cost at 0.30 GigaFLOPs and reaches 66.15\% accuracy. EfficientNetB3 offers a balanced result with 87.42\% accuracy at 1.83 GigaFLOPs.


\begin{figure*}[!ht]
    \centering

    \begin{subfigure}[b]{0.8\textwidth}
        \centering
        \includegraphics[width=\linewidth]{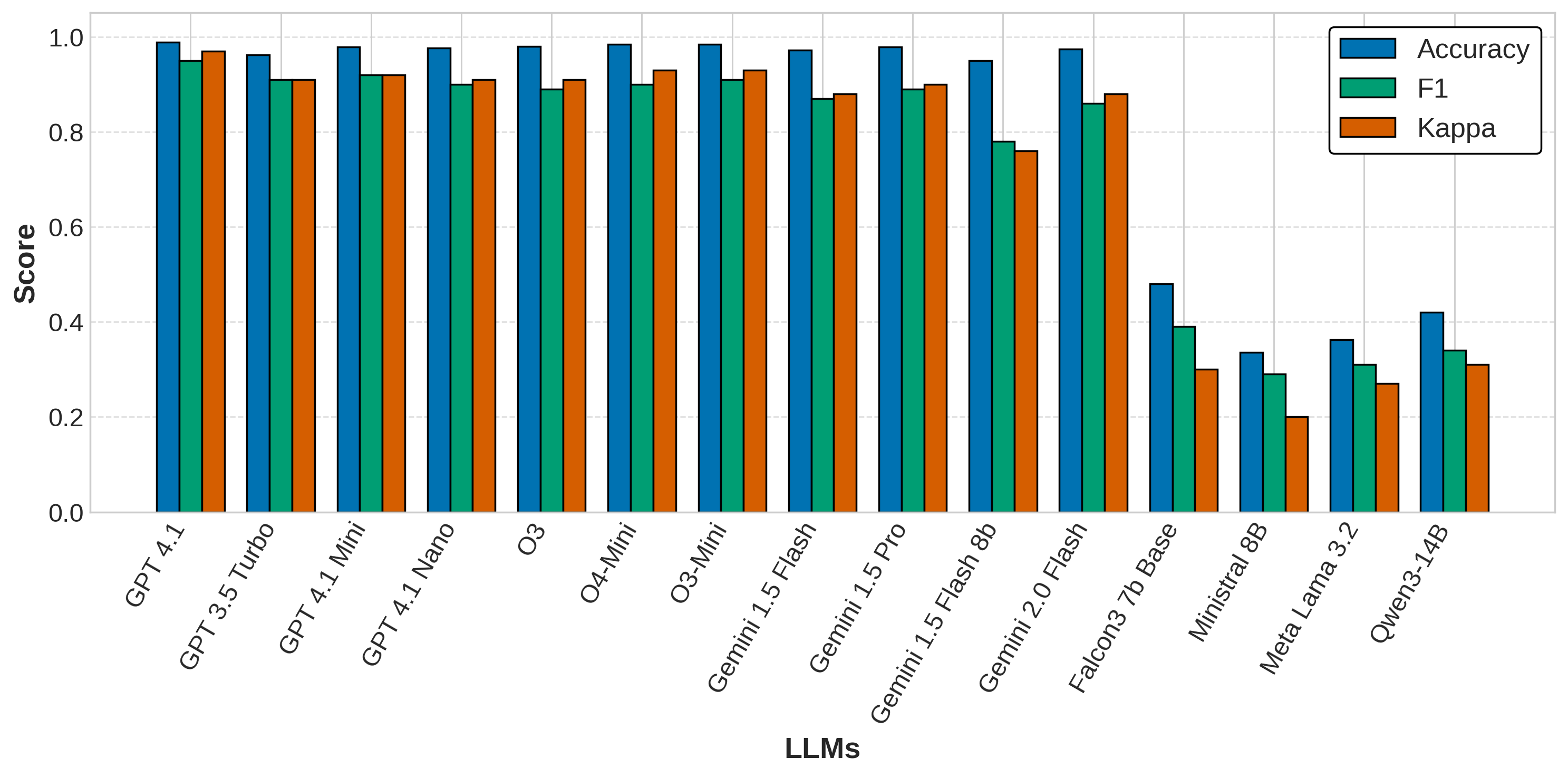}
        \caption{Performance evaluation on the Level 1 multiple-choice task prompt}
        \label{Proposed_Model_metrics_p1}
    \end{subfigure}
    \par\vspace{0.5cm} 

    \begin{subfigure}[b]{0.8\textwidth}
        \centering
        \includegraphics[width=\linewidth]{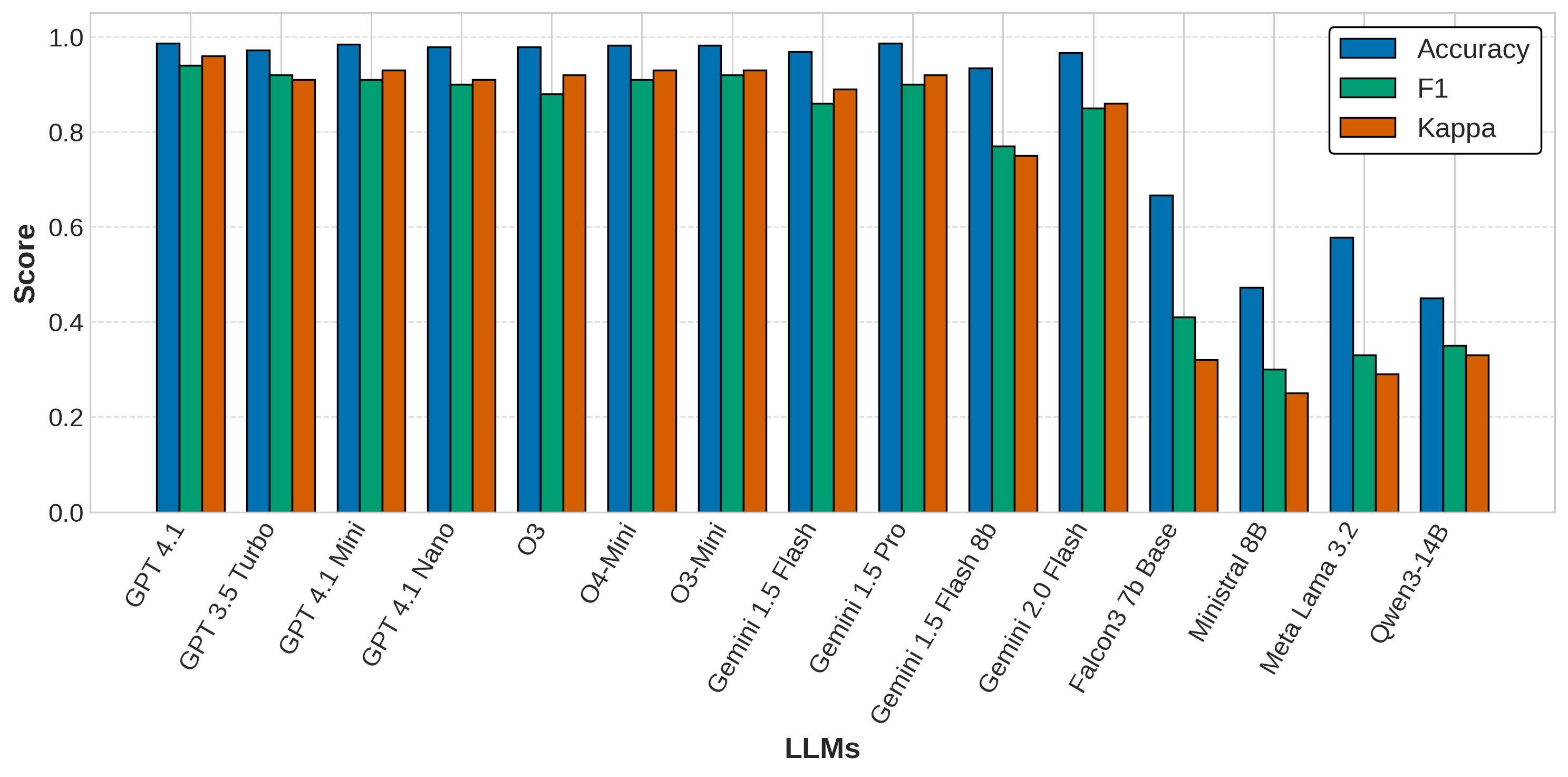}
        \caption{Performance evaluation on the Level 2 multiple-choice task prompt}
        \label{Proposed_Model_metrics_p2}
    \end{subfigure}
    \par\vspace{0.5cm}

    \begin{subfigure}[b]{0.8\textwidth}
        \centering
        \includegraphics[width=\linewidth]{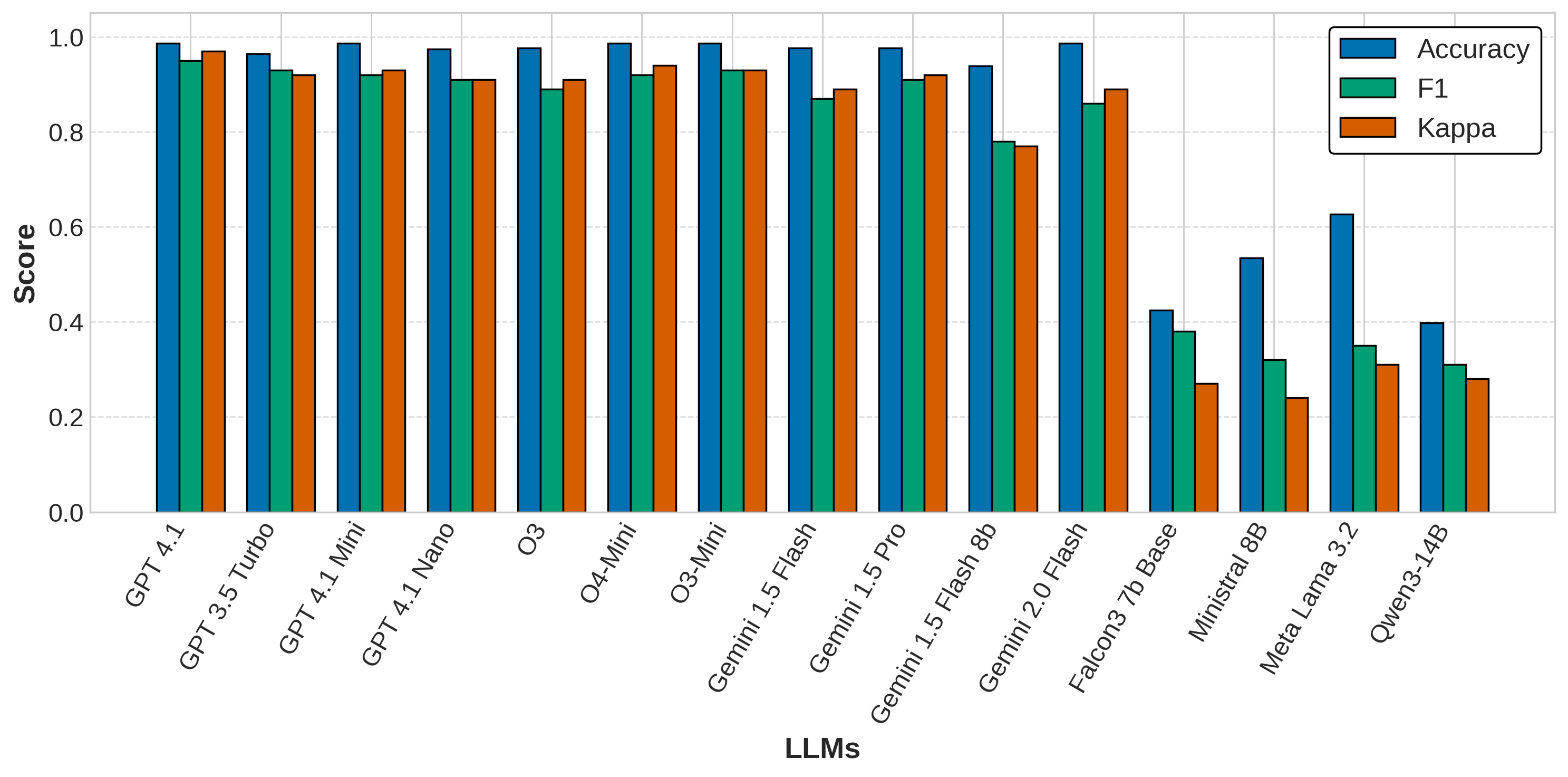}
        \caption{Performance evaluation on the Level 3 multiple-choice task prompt}
        \label{Proposed_Model_metrics_p3}
    \end{subfigure}

    \label{Proposed_Model_metrics_all}
\end{figure*}


\begin{figure*}[!ht]
    \centering
    \ContinuedFloat 

    \begin{subfigure}[b]{0.8\textwidth}
        \centering
        \includegraphics[width=\linewidth]{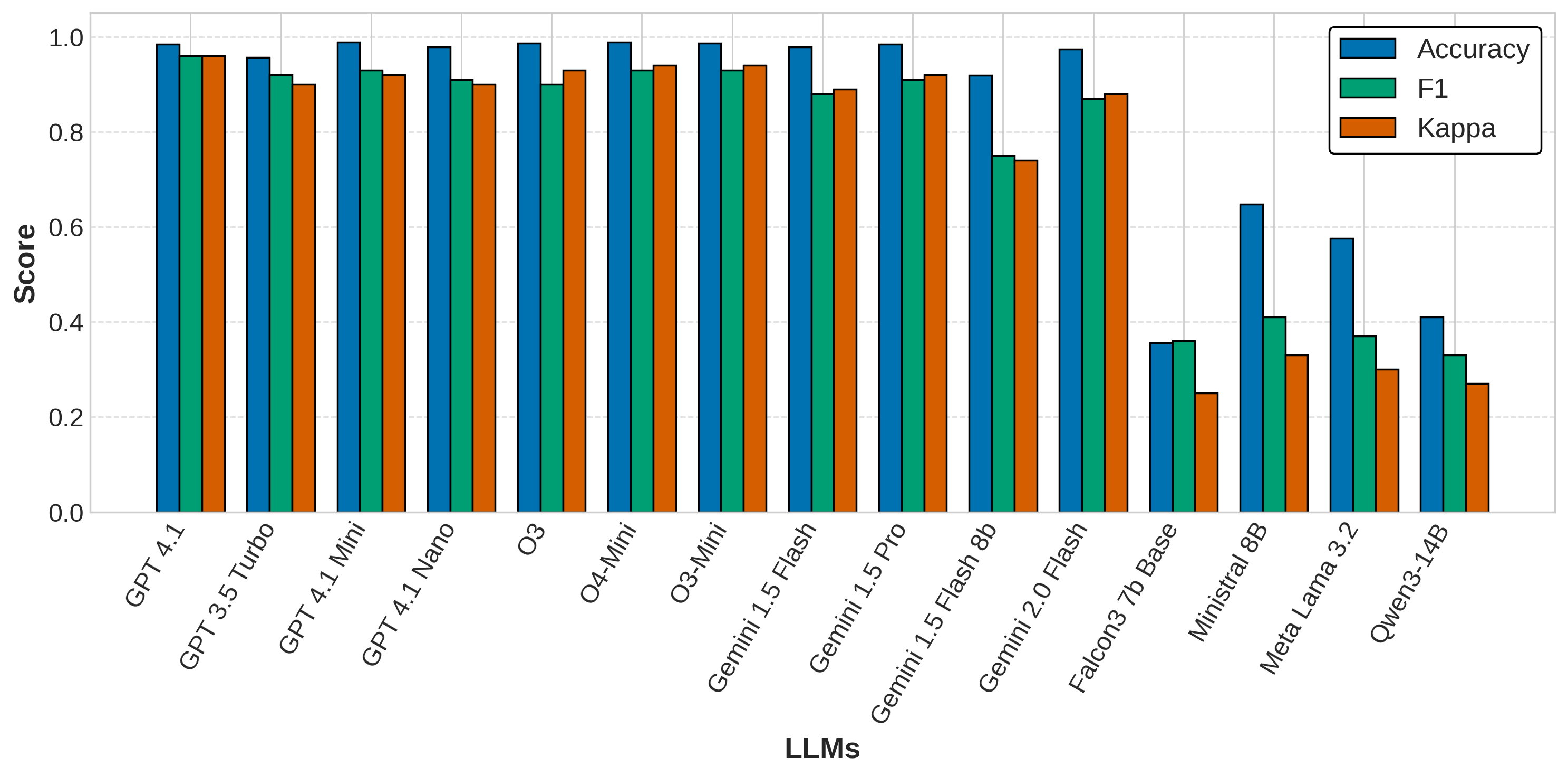}
        \caption{Performance evaluation on the Level 4 multiple-choice task prompt}
        \label{Proposed_Model_metrics_p4}
    \end{subfigure}

    \caption{Performance evaluation of fifteen best LLMs with four different levels of prompts for the multiple-choice task.}
\end{figure*}

\subsection{Performance Evaluation for LLM-based Multi-step Reasoning}
\label{Analysis of best performing LLMs}

    \subsubsection{LLM selection}
  
    Thirty two model evaluated on GastroLLM-Bench using BERTScore, Cosine Similarity, and ROUGE-L at this stage. Based on these results, 15 models were chosen for further experiments. GPT-4.1, GPT-3.5-Turbo, GPT-4.1-Nano, GPT-4.1-Mini, O3, O4-Mini, O3-Mini, and Gemini-1.5-Flash showed consistently strong performance across all disease classes (see \autoref{fig:llm_selection_composite_score_graph}).
    
    To identify the best-performing models, we created a composite score by normalizing BERTScore, cosine similarity, and ROUGE-L, and then averaging the normalized values. This provided a balanced measure of semantic alignment, contextual consistency, and overall reasoning quality. The results reveal a clear pattern that commercial models currently hold a noticeable advantage in medical reasoning tasks. Models with more than seven billion parameters also tended to perform better. This experiment suggests that larger parameter counts still play an important role in capturing clinical nuance and producing stable reasoning. The strongest results were observed across seven major model families. These included four models from OpenAI’s GPT series, three from OpenAI’s reasoning-oriented models, four from the Gemini family, and one model each from the Qwen, LLaMA, Mistral, and Falcon families. This range shows that high performance is not restricted to a single architecture, but it also highlights that only a small portion of the broader LLM ecosystem is capable of generating consistent, medically grounded explanations.

\begin{figure*}[!ht]
    \centering

    \begin{subfigure}[b]{0.8\textwidth} 
        \centering
        \includegraphics[width=\linewidth]{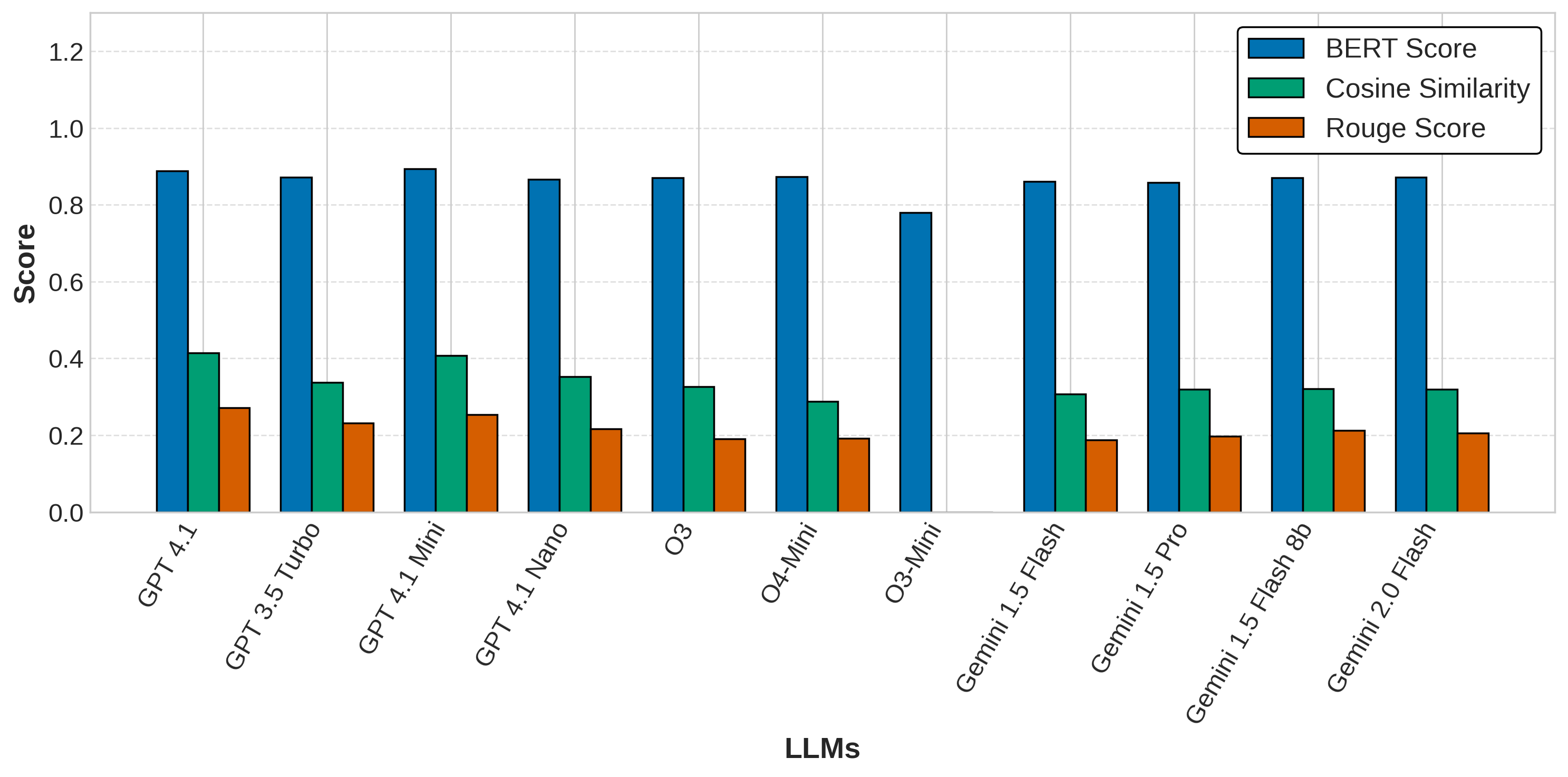}
        \caption{Performance evaluation on the Level 1 reasoning task prompt}
        \label{Proposed_Model_confusion_matrix_p1}
    \end{subfigure}
    \par\vspace{0.2cm}

    \begin{subfigure}[b]{0.8\textwidth}
        \centering
        \includegraphics[width=\linewidth]{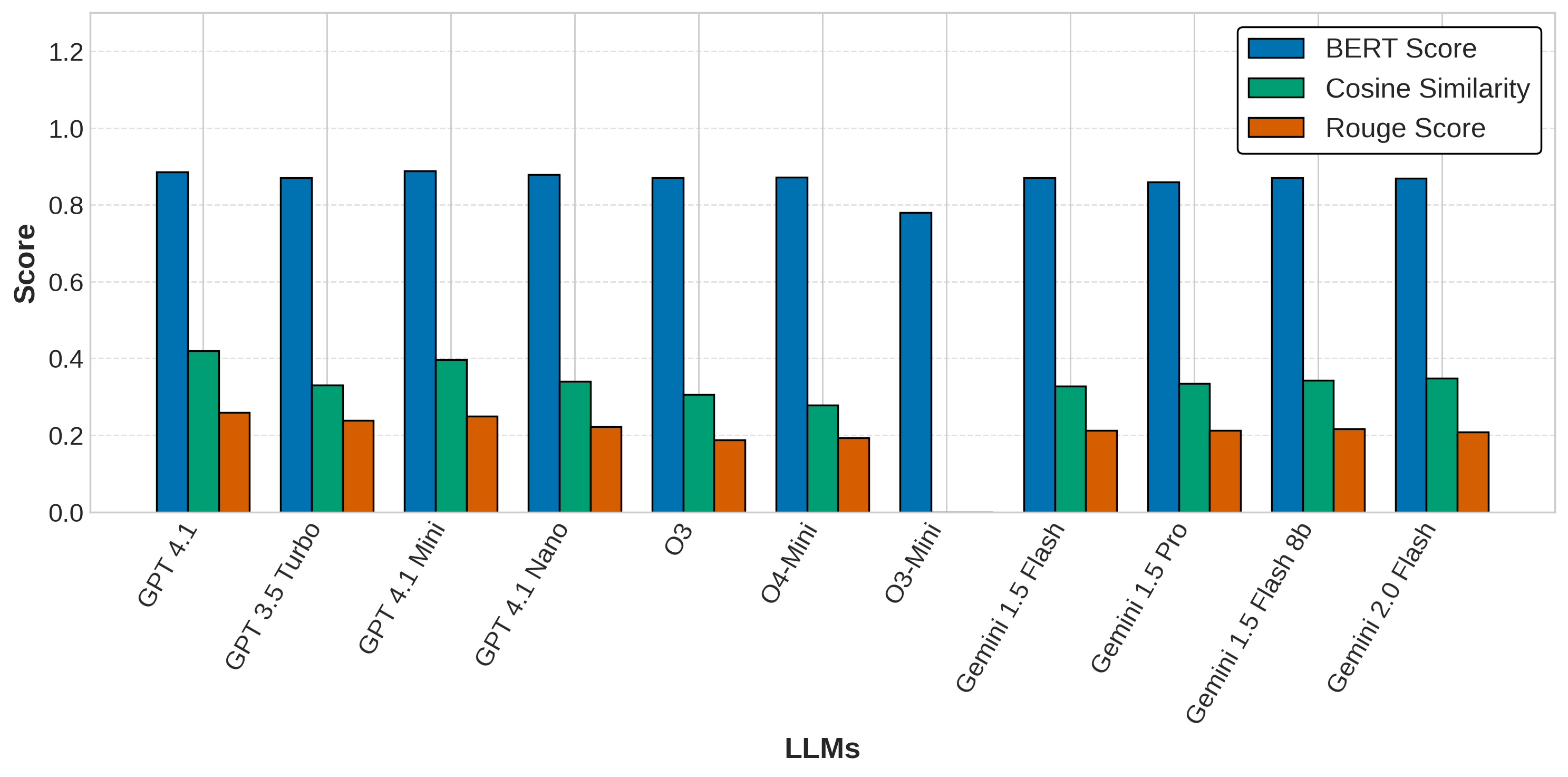}
        \caption{Performance evaluation on the Level 2 reasoning task prompt}
        \label{Proposed_Model_confusion_matrix_p2}
    \end{subfigure}
    \par\vspace{0.2cm}

    \begin{subfigure}[b]{0.8\textwidth}
        \centering
        \includegraphics[width=\linewidth]{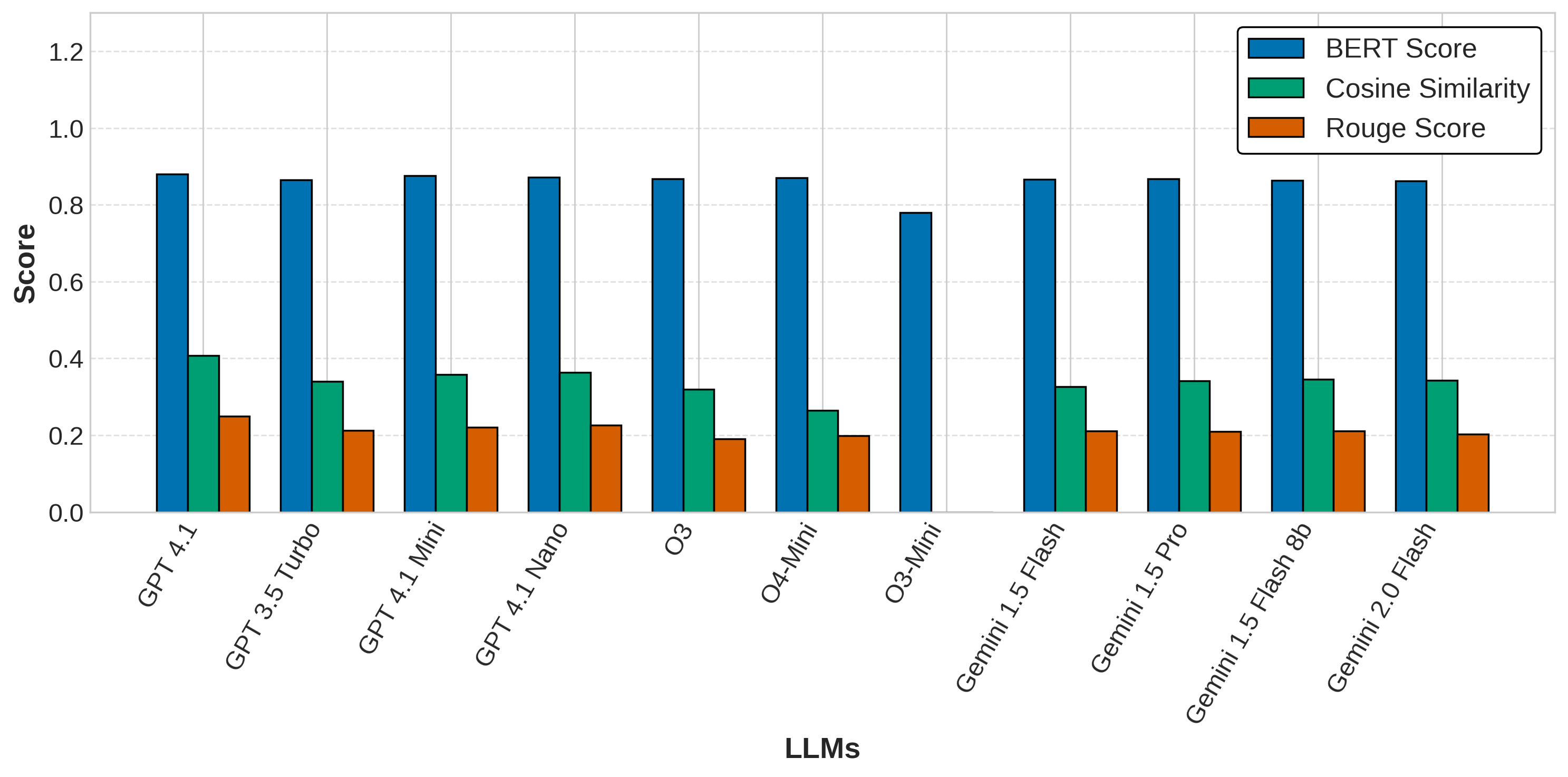}
        \caption{Performance evaluation on the Level 3 reasoning task prompt}
        \label{Proposed_Model_confusion_matrix_p3}
    \end{subfigure}

    \label{Proposed_Model_confusion_matrix_all}
\end{figure*}


\begin{figure*}[!ht]
    \centering
    \ContinuedFloat 

    \begin{subfigure}[b]{0.8\textwidth}
        \centering
        \includegraphics[width=\linewidth]{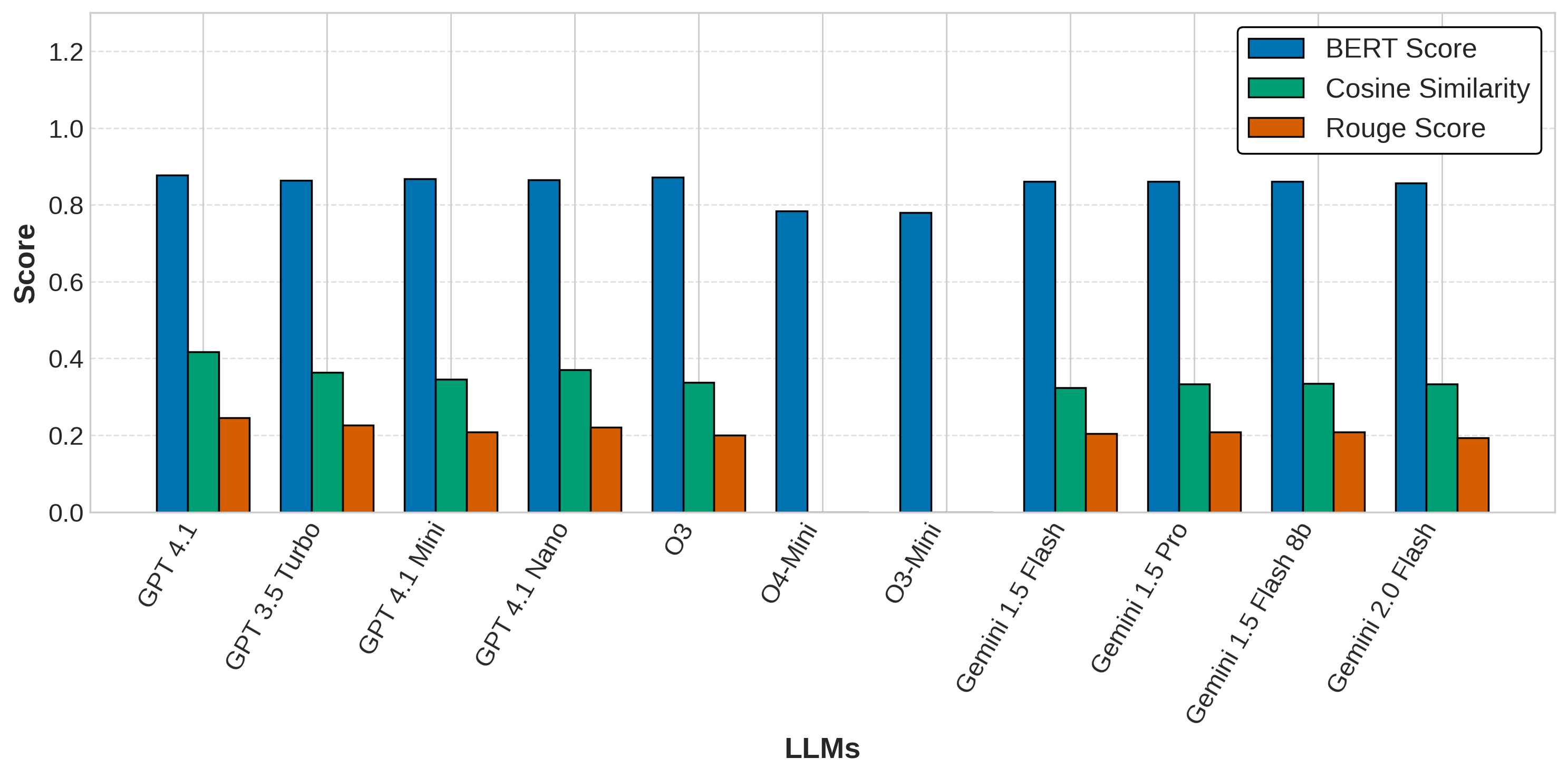}
        \caption{Performance evaluation on the Level 4 reasoning task prompt}
        \label{Proposed_Model_confusion_matrix_p4}
    \end{subfigure}

    \caption{Performance evaluation of fifteen best LLMs with four different levels of prompts for the multiple-step reasoning task.}
\end{figure*}

\begin{sidewaystable*}[ht]
\centering
\small
\caption{Reasoning performance of fifteen LLMs across four levels of prompts}
\label{table:llm_performance_reasoning}
\begin{tabular}{llrrrrrrrrrrrr}
    \toprule
    \multirow{2}{*}{Model} & \multirow{2}{*}{Class} &
    \multicolumn{3}{c}{Prompt 1} &
    \multicolumn{3}{c}{Prompt 2} &
    \multicolumn{3}{c}{Prompt 3} &
    \multicolumn{3}{c}{Prompt 4} \\
    & & 
    BERT & Cosine & Rouge &
    BERT & Cosine & Rouge &
    BERT & Cosine & Rouge &
    BERT & Cosine & Rouge \\
    \midrule
   \multirow{5}{*}{GPT 4.1} & Polyps & 0.9006 & 0.4117 & 0.2840 & 0.8919 & 0.4055 & 0.2535 & 0.8962 & 0.4053 & 0.2690 & 0.8944 & 0.4095 & 0.2620 \\
    & Ulcerative Colitis & 0.8968 & 0.4232 & 0.2698 & 0.8894 & 0.4136 & 0.2487 & 0.8923 & 0.4155 &	0.2639 & 0.8855 & 0.4047 & 0.2342 \\
    & Esophagitis & 0.8951 & 0.4041 & 0.2620 & 0.8805 & 0.3599 & 0.2182 & 0.8907 & 0.3918 & 0.2503 & 0.8848 & 0.3822 & 0.2299 \\
    & Healthy Class & 0.8818 & 0.5269 & 0.2926 & 0.8836 & 0.5546 & 0.3092 & 0.8593 & 0.4630 & 0.2416 & 0.8666 & 0.4901 & 0.2467 \\
    & Removal of Lesions & 0.8736 & 0.3975 & 0.2407 & 0.8779 & 0.3989 & 0.2391 & 0.8652 & 0.3752 & 0.2254 & 0.8768 & 0.4109 & 0.2548 \\
    \hline
    
    \multirow{5}{*}{GPT 3.5 turbo} & Polyps & 0.8923 & 0.3364 & 0.2512 & 0.8994 & 0.3700 & 0.2749 & 0.8656 & 0.3074 & 0.1935 & 0.8778 & 0.3481 & 0.2346 \\
    & Ulcerative Colitis & 0.8958 & 0.3921 & 0.2678 & 0.8963 & 0.3875 & 0.2718 & 0.8642 & 0.3249 & 0.1942 & 0.8776 & 0.3619 & 0.2350 \\
    & Esophagitis & 0.8968 & 0.3772 & 0.2742 & 0.8968 & 0.3782 & 0.2634 & 0.8718 & 0.3391 & 0.2141 & 0.8827 & 0.3681 & 0.2374 \\
    & Healthy Class & 0.8211 & 0.1904 & 0.1092 & 0.8265 & 0.2057 & 0.1125 & 0.8694 & 0.4874 & 0.2472 & 0.8609 & 0.4435 & 0.2252 \\
    & Removal of Lesions & 0.8564 & 0.2867 & 0.1909 & 0.8399 & 0.2140 & 0.1361 & 0.8639 & 0.3435 & 0.2175 & 0.8616 & 0.3639 & 0.2239 \\
    \hline
     
    \multirow{5}{*}{GPT 4.1 Mini} & Polyps & 0.8944 & 0.4130 & 0.2751 & 0.8918 & 0.3998 & 0.2726 & 0.8731 & 0.3335 & 0.2263 & 0.8615 & 0.3034 & 0.1902 \\
    & Ulcerative Colitis & 0.8944 & 0.4130 & 0.2751 & 0.8918 & 0.3998 & 0.2726 & 0.8731 & 0.3335 & 0.2263 & 0.8615 & 0.3034 & 0.1902 \\
    & Esophagitis & 0.9065 & 0.4016 & 0.2600 & 0.9004 & 0.3885 & 0.2405 & 0.8738 & 0.3137 & 0.2050 & 0.8661 & 0.3420 & 0.1945 \\
    & Healthy Class & 0.8802 & 0.4782 & 0.2603 & 0.8868 & 0.4869 & 0.2738 & 0.8721 & 0.4390 & 0.2371 & 0.8565 & 0.4879 & 0.2125 \\
    & Removal of Lesions & 0.8890 & 0.4020 & 0.2459 & 0.8895 & 0.3945 & 0.2394 & 0.8712 & 0.3640 & 0.2191 & 0.8722 & 0.3765 & 0.2332 \\
    \hline
      
    \multirow{5}{*}{GPT 4.1 Nano} & Polyps & 0.8737 & 0.3278 & 0.2141 & 0.8880 & 0.3780 & 0.2508 & 0.8819 & 0.3599 & 0.2296 & 0.8655 & 0.3186 & 0.1960 \\
    & Ulcerative Colitis & 0.8674 & 0.3124 & 0.1886 & 0.8926 & 0.3812 & 0.2436 & 0.8844 & 0.3817 & 0.2285 & 0.8725 & 0.3641 & 0.2070 \\
    & Esophagitis & 0.8819 & 0.3394 & 0.2210 & 0.8948 & 0.3704 & 0.2405 & 0.8908 & 0.3915 & 0.2388 & 0.8791 & 0.3897 & 0.2222 \\
    & Healthy Class & 0.8330 & 0.3323 & 0.1563 & 0.8440 & 0.3470 & 0.1693 & 0.8661 & 0.4686 & 0.2395 & 0.8644 & 0.5070 & 0.2320 \\
    & Removal of Lesions & 0.8229 & 0.1735 & 0.0853 & 0.8292 & 0.2004 & 0.1011 & 0.8572 & 0.3416 & 0.1986 & 0.8611 & 0.3724 & 0.2072 \\
    \hline
      
    \multirow{5}{*}{O3} & Polyps & 0.8688 & 0.3231 & 0.2021 & 0.8665 & 0.3136 & 0.1936 & 0.8649 & 0.3034 & 0.1897 & 0.8702 & 0.3221 & 0.2040 \\
    & Ulcerative Colitis & 0.8777 & 0.3115 & 0.1888 & 0.8742 & 0.3040 & 0.1855 & 0.8705 & 0.2928 & 0.1799 & 0.8780 & 0.3029 & 0.1831 \\
    & Esophagitis & 0.8826 & 0.3097 & 0.1985 & 0.8746 & 0.3002 & 0.1855 & 0.8780 & 0.3060 & 0.1884 & 0.8818 & 0.3070 & 0.1991 \\
    & Healthy Class & 0.8558 & 0.3898 & 0.1913 & 0.8621 & 0.4006 & 0.2004 & 0.8555 & 0.3895 & 0.1965 & 0.8639 & 0.4251 & 0.2169 \\
    & Removal of Lesions & 0.8635 & 0.2930 & 0.1665 & 0.8698 & 0.3191 & 0.1736 & 0.8655 & 0.3213 & 0.1816 & 0.8737 & 0.3357 & 0.1975 \\
    \hline

    \multirow{5}{*}{O4-Mini} & Polyps & 0.8735 & 0.2923 & 0.2148 & 0.8733 & 0.2890 & 0.2168 & 0.8650 & 0.2659 & 0.1993 & 0.7817 & 0.0000 & 0.0000 \\
    & Ulcerative Colitis & 0.8852 & 0.2766 & 0.2077 & 0.8825 & 0.2823 & 0.1980 & 0.8749 & 0.2540 & 0.1896 & 0.7791 & 0.0000 & 0.0000 \\
    & Esophagitis & 0.8832 & 0.2654 & 0.2031 & 0.8845 & 0.2832 & 0.2047 & 0.8780 & 0.2639 & 0.1961 & 0.8832 & 0.2790 & 0.2034 \\
    & Healthy Class & 0.8608 & 0.3267 & 0.1911 & 0.8650 & 0.3525 & 0.1976 & 0.8679 & 0.3704 & 0.2119 & 0.7779 & 0.0000 & 0.0000 \\
    & Removal of Lesions & 0.8652 & 0.2690 & 0.1821 & 0.8614 & 0.2686 & 0.1709 & 0.8669 & 0.2705 & 0.1905 & 0.7842 & 0.0000 & 0.0000 \\
    
\bottomrule
\end{tabular}
\end{sidewaystable*}

\begin{sidewaystable*}[t]
\small
\centering
\caption{Reasoning performance of fifteen LLMs across four levels of prompts (continued)}
\label{table:llm_performance_reasoning_2}
\begin{tabular}{llrrrrrrrrrrrr}
    \toprule
    \multirow{2}{*}{Model} & \multirow{2}{*}{Class} &
    \multicolumn{3}{c}{Prompt 1} &
    \multicolumn{3}{c}{Prompt 2} &
    \multicolumn{3}{c}{Prompt 3} &
    \multicolumn{3}{c}{Prompt 4} \\
    & & 
    BERT & Cosine & Rouge &
    BERT & Cosine & Rouge &
    BERT & Cosine & Rouge &
    BERT & Cosine & Rouge \\
    \midrule
    
    \multirow{5}{*}{O3-Mini} & Polyps & 0.7817 & 0.0000 & 0.0000 & 0.7817 & 0.0000 & 0.0000 & 0.7817 & 0.0000 & 0.0000 & 0.7817 & 0.0000 & 0.0000 \\
    & Ulcerative Colitis & 0.7791 & 0.0000 & 0.0000 & 0.7791 & 0.0000 & 0.0000 & 0.7791 & 0.0000 & 0.0000 & 0.7791 & 0.0000 & 0.0000 \\
    & Esophagitis & 0.7763 & 0.0000 & 0.0000 & 0.7763 & 0.0000 & 0.0000 & 0.7763 & 0.0000 & 0.0000 & 0.7763 & 0.0000 & 0.0000 \\
    & Healthy Class & 0.7779 & 0.0000 & 0.0000 & 0.7779 & 0.0000 & 0.0000 & 0.7779 & 0.0000 & 0.0000 & 0.7779 & 0.0000 & 0.0000 \\
    & Removal of Lesions & 0.7842 & 0.0000 & 0.0000 & 0.7842 & 0.0000 & 0.0000 & 0.7842 & 0.0000 & 0.0000 & 0.7842 & 0.0000 & 0.0000 \\
    \hline

    \multirow{5}{*}{Gemini 1.5 Flash} & Polyps & 0.8588 & 0.2684 & 0.1887 & 0.8711 & 0.3242 & 0.2212 & 0.8706 & 0.3240 & 0.2121 & 0.8660 & 0.2951 & 0.1989 \\
    & Ulcerative Colitis & 0.8620 & 0.2655 & 0.1747 & 0.8839 & 0.3552 & 0.2226 & 0.8837 & 0.3629 & 0.2285 & 0.8882 & 0.3685 & 0.2358 \\
    & Esophagitis & 0.8584 & 0.2584 & 0.1705 & 0.8836 & 0.3332 & 0.2163 & 0.8834 & 0.3307 & 0.2203 & 0.8861 & 0.3480 & 0.2280 \\
    & Healthy Class & 0.8294 & 0.3071 & 0.1516 & 0.8340 & 0.3169 & 0.1538 & 0.8348 & 0.3667 & 0.1705 & 0.8556 & 0.4248 & 0.2110 \\
    & Removal of Lesions & 0.8424 & 0.2702 & 0.1509 & 0.8444 & 0.2904 & 0.1515 & 0.8666 & 0.3290 & 0.2017 & 0.8622 & 0.3348 & 0.2093 \\
    \hline

    \multirow{5}{*}{Gemini 1.5 Pro} & Polyps & 0.8424 & 0.2702 & 0.1509 & 0.8444 & 0.2904 & 0.1515 & 0.8666 & 0.3290 & 0.2017 & 0.8622 & 0.3348 & 0.2093 \\
    & Ulcerative Colitis & 0.8775 & 0.3597 & 0.2227 & 0.8752 & 0.3685 & 0.2067 & 0.8865 & 0.4078 & 0.2343 & 0.8793 & 0.3572 & 0.2191 \\
    & Esophagitis & 0.8705 & 0.3208 & 0.1944 & 0.8714 & 0.3312 & 0.2041 & 0.8874 & 0.3968 & 0.2368 & 0.8839 & 0.3636 & 0.2280 \\
    & Healthy Class & 0.8705 & 0.3208 & 0.1944 & 0.8714 & 0.3312 & 0.2041 & 0.8874 & 0.3968 & 0.2368 & 0.8839 & 0.3636 & 0.2280 \\
    & Removal of Lesions & 0.8412 & 0.2411 & 0.1361 & 0.8381 & 0.2107 & 0.1203 & 0.8541 & 0.3678 & 0.1850 & 0.8652 & 0.3225 & 0.2086 \\
    \hline

    \multirow{5}{*}{Gemini 1.5 Flash 8b} & Polyps & 0.8811 & 0.3011 & 0.2323 & 0.8795 & 0.3114 & 0.2274 & 0.8624 & 0.3067 & 0.1955 & 0.8657 & 0.3063 & 0.2051 \\
    & Ulcerative Colitis & 0.8860 & 0.3300 & 0.2269 & 0.8893 & 0.3647 & 0.2248 & 0.8770 & 0.3543 & 0.2108 & 0.8834 & 0.3673 & 0.2299 \\
    & Esophagitis & 0.8826 & 0.2938 & 0.2137 & 0.8851 & 0.3267 & 0.2161 & 0.8723 & 0.3132 & 0.1999 & 0.8778 & 0.3397 & 0.2102 \\
    & Healthy Class & 0.8460 & 0.3685 & 0.1860 & 0.8441 & 0.4101 & 0.1881 & 0.8542 & 0.4170 & 0.2126 & 0.8626 & 0.4309 & 0.2388 \\
    & Removal of Lesions & 0.8551 & 0.2980 & 0.1731 & 0.8548 & 0.3007 & 0.1758 & 0.8717 & 0.3602 & 0.2092 & 0.8723 & 0.3702 & 0.2178 \\
    \hline

    \multirow{5}{*}{Gemini 2.0 Flash} & Polyps & 0.8815 & 0.3204 & 0.2351 & 0.8732 & 0.3216 & 0.2143 & 0.8626 & 0.3170 & 0.2015 & 0.8512 & 0.2769 & 0.1804 \\
    & Ulcerative Colitis & 0.8836 & 0.3604 & 0.2396 & 0.8849 & 0.3811 & 0.2325 & 0.8754 & 0.3817 & 0.2140 & 0.8541 & 0.2862 & 0.1805 \\
    & Esophagitis & 0.8785 & 0.3027 & 0.2085 & 0.8841 & 0.3609 & 0.2259 & 0.8700 & 0.3477 & 0.2060 & 0.8547 & 0.2814 & 0.1778 \\
    & Healthy Class & 0.8342 & 0.3122 & 0.1667 & 0.8333 & 0.3084 & 0.1607 & 0.8453 & 0.4189 & 0.2101 & 0.8423 & 0.4014 & 0.2009 \\
    & Removal of Lesions & 0.8458 & 0.2451 & 0.1526 & 0.8553 & 0.3485 & 0.1795 & 0.8578 & 0.3491 & 0.2035 & 0.8582 & 0.3599 & 0.2098 \\
\bottomrule
\end{tabular}
\end{sidewaystable*}

    \subsubsection{Multiple Choice Task}
    In the classification stage, the LLMs were evaluated based on their responses to multiple-choice questions. The results were assessed using Accuracy, F1 Score, and Cohen’s Kappa metrics (\autoref{table:llm_performance_classification} and \ref{table:llm_performance_classification_2}).
    
    \textbf{(a) Small Model (less than 10 billion parameters):}
    Among the small models, Gemini-1.5-Flash, Gemini-2.0-Flash, GPT-4.1-Mini, GPT-4.1-Nano, and Gemini-1.5-Flash (8B) performed strongly, with accuracy, F1 Score, and Cohen’s kappa all above 85\%. For an LLM, achieving high scores across these metrics indicates more than correct classification—it reflects better contextual understanding, stronger semantic alignment with clinical descriptions, and more consistent reasoning across samples. In contrast, Mistral-8B and Falcon-3 7B Base stayed below 50\% on all three metrics, suggesting weaker domain grounding, frequent confusion between clinically similar classes, and limited agreement with expected reasoning patterns. Since both are open-source, they still have room for improvement through targeted fine-tuning. Performance on the Polyp class showed greater variation. Gemini-1.5-Flash reached roughly 91\% accuracy with an F1 score of 80\% and a Cohen’s kappa of 84\%, indicating strong sensitivity to the features associated with polyp-related cases. Gemini-1.5-Flash (8B) achieved around 80\% accuracy but dropped to an F1 score of 57\% and a kappa of 53\%, suggesting difficulty with borderline cases and reduced stability when distinguishing this class from similar conditions. Even so, both models performed well across the remaining disease classes, implying that these weaknesses are limited to specific categories. All five proprietary models remained stable across different prompts, showing deeper domain understanding, stronger semantic robustness, and less dependence on prompt wording. Among the weaker models, Mistral-8B produced its best results under Prompt 4, while Falcon-3 7B Base performed better under Prompt 2 but declined under Prompt 4, highlighting their sensitivity to prompt formulation and their limited ability to generalize reasoning across varying prompting conditions.
    
    \textbf{(b) Mid-Size Models (Parameter size \( > 10B \) and \( < 50B \)):}
    Among the four mid-size models, GPT-4.1-Mini and GPT-3.5-Turbo performed strongly, scoring above 85\% on all evaluation metrics. Their higher scores suggest better contextual understanding and more stable alignment with the clinical reasoning patterns expected in this task. In contrast, LLaMA-3.2 (11B) and Qwen-3 (14B) showed weaker performance, with all metrics falling below 60\%. This indicates limitations in their ability to capture medical nuance, maintain semantic consistency, or generalize across different reasoning prompts. Both lower-performing models are open-source, making them strong candidates for domain-specific fine-tuning to address these gaps and improve their reasoning quality.
    
    \textbf{(c) Large Models (Parameter size \( > 50B \)):}
    All four large models - O3-Mini, Gemini-1.5-Pro, GPT-4.1, and O3, performed strongly, with scores above 90\% across all evaluation metrics. Although their performance dipped slightly for the Esophagitis, Polyp, and Ulcerative Colitis classes, they achieved perfect scores for the Healthy and Removal of Lesions classes, showing strong alignment with cases that present clearer visual and clinical cues. All four models remained stable across most prompt variations, though a small decline appeared under Prompt 3, suggesting that certain forms of reasoning still introduce mild instability even in high-capacity models.

    When comparing all 15 models, the top five performers were O3-Mini, with 99\% accuracy, an F1 score of 94\%, and a Cohen’s kappa of 96\%; Gemini-1.5-Pro, with 98\% accuracy, an F1 of 95\%, and a kappa of 95\%; O4-Mini, with 98\% accuracy, an F1 of 92\%, and a kappa of 95\%; GPT-4.1, with 99\% accuracy, an F1 of 90\%, and a kappa of 97\%; and O3, with 98\% accuracy, an F1 of 91\%, and a kappa of 96\%. Four of these five models exceed 200 billion parameters, reinforcing the value of large-scale architectures for clinically grounded reasoning. At this stage of the analysis, only models scoring above 80\% on all three metrics were retained for the reasoning-label evaluation. Four models—Mistral-8B, Falcon-3 7B Base, Qwen-3 14B, and Llama-3.2 11B—did not meet this threshold. Notably, all four are open-source models, highlighting a clear performance gap between proprietary and open-source systems in medical reasoning tasks, likely driven by differences in training data quality, model scale, and domain exposure.

    
    

\subsubsection{Reasoning Generation Task}

\label{Reasoning Genaration Task}
    Three metrics were used to evaluate the reasoning outputs: BERTScore, cosine similarity, and ROUGE-L. In the BERTScore analysis, 10 of the 11 models performed well. GPT-4.1 and GPT-4.1 Mini led with scores of 88\%, followed closely by GPT-3.5 Turbo, Gemini 1.5 Flash 8B, O3, and GPT-4.1 Nano, each scoring 87\%. Gemini 1.5 Pro, Gemini 1.5 Flash, and Gemini 2.0 Flash showed slightly lower but still strong scores of 86\%, while O4-Mini reached 85\%. The weakest performance came from O3-Mini, which achieved 78\%.

For cosine similarity, GPT-4.1, GPT-4.1 Mini, and GPT-4.1 Nano again ranked highest with scores of 42\%, 39\%, and 35\%. Gemini 1.5 Flash 8B, GPT-3.5 Turbo, and Gemini 1.5 Pro each obtained 0.34. Gemini 2.0 Flash and O3 scored 33\%, followed by Gemini 1.5 Flash with 32\%. O4-Mini dropped to 23\%, and O3-Mini recorded the lowest value of 0.00. The ROUGE-L results followed a similar pattern. GPT-4.1 scored highest at 25.5\%, with GPT-4.1 Mini next at 23.6\%. GPT-3.5 Turbo reached 21.9\%, while Gemini 1.5 Flash 8B and GPT-4.1 Nano scored 0.210 and 0.203. Gemini 2.0 Flash and Gemini 1.5 Pro produced comparable values of 20\% and 19.8\%, and Gemini 1.5 Flash and O3 scored 19.6\% and 19.1\%. O4-Mini dropped to 15.9\%, while O3-Mini again showed no measurable performance with 0.000.

\begin{figure*}[!htb]
    \centering
    \includegraphics[width=0.5\linewidth]{
        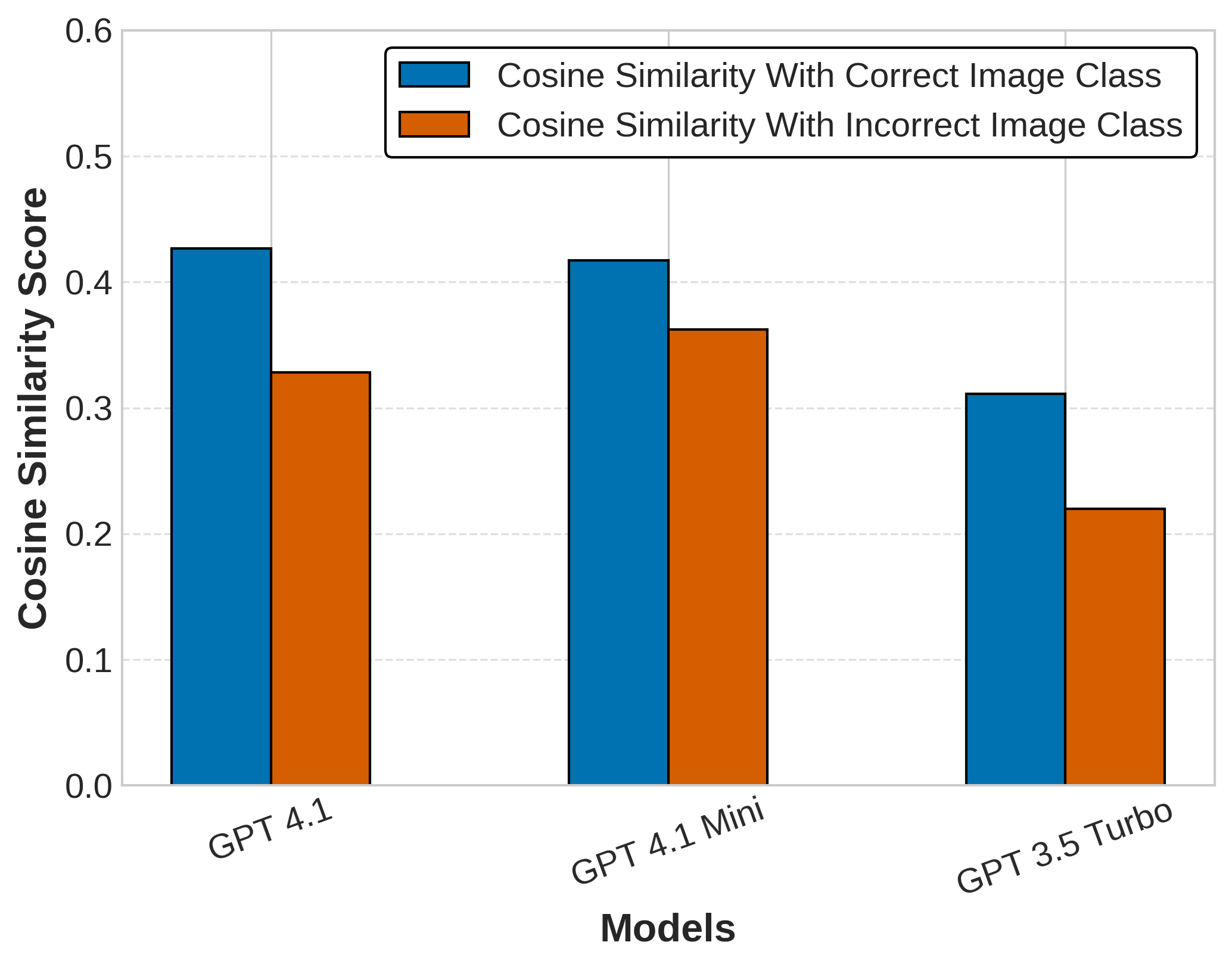
    }
    \caption{Performance comparison for multi-step reasoning generation of the best-performing LLMs in two different settings, when the image is predicted correctly and when the image is mispredicted from the proposed MobileCoAtNet (DL model).}
    \label{fig:Cosine_Similarity_Comparison}
\end{figure*} 

Overall, GPT-4.1 and GPT-4.1 Mini consistently ranked at the top across all three metrics. Although these LLMs performed the highest for correct image labels (when the DL model predicts an image class correctly), they fall short for incorrect image classes refer to \autoref{fig:Cosine_Similarity_Comparison}. Again, the strongest models show noticeable gaps in semantic alignment, stability, and contextual consistency, indicating that none of them can yet be trusted to produce reasoning outputs without human oversight.

\subsubsection{Alignment of Top-Performing LLMs with Expert-Annotated Reasoning}
\label{Analysis of the best performance LLM with human expertise}

    As the results in \autoref{table:llm_reasoning_leaderboard_reversed} show, no model achieved the `High-CRA' status. The top models, \textbf{GPT 4.1} (53\%) and \textbf{GPT 4.1 Mini} (51.5\%), fall into the `Moderate-CRA' category, indicating they provide competent reasoning, especially on general prompts. The remaining models, including the \textbf{Gemini series} and \textbf{GPT 3.5 turbo}, are now categorized as `Low-CRA' under this strict threshold, as their ability to reason with minimal context falls below the 50\% cutoff.

\section{Discussion}
\label{discussion}

In our study we show that combining deep-learning–based disease classification with LLM-driven reasoning helps narrow the gap between image recognition and clinical explanation, but it also exposes several limitations. For $RQ_1$, our results show that the quality of classification directly shapes the quality of reasoning. Because LLMs cannot interpret images on their own and rely entirely on the predicted class label, a correct prediction leads to meaningful clinical reasoning, while an incorrect prediction immediately results in flawed explanations. This makes upstream classification accuracy essential. For $RQ_2$, we find that LLMs are capable of producing useful, multi-step clinical reasoning when the DL model’s prediction is correct. They can describe causes, symptoms, treatment options, dietary suggestions, and follow-up care. However, their performance varies across model sizes: larger models tend to be more consistent and detailed, while smaller models depend heavily on prompt wording and often miss important medical details. For $RQ_3$, our findings indicate that no current LLM produces reasoning that matches expert-level judgment. Although future models may come closer to expert reasoning, they would still require supervision before being used in sensitive medical environments. Models exhibiting moderate clinical reasoning alignment will need domain adaptation or fine-tuning to be reliable, while less consistent models will require substantial retraining with structured datasets before they can be trusted. Finally, the study demonstrates that while the \textbf{DL\textsuperscript{3}M} pipeline improves interpretability and clinical alignment compared to using either model alone, its reliability still depends on accurate classification, selecting appropriate LLMs, careful prompt design, and continuous human oversight - especially in medical settings where errors can have serious consequences.

\section{Conclusion}
\label{conclusion}
From our experiment, we present a complete view, how DL–based disease classification and LLMs can work together to support clinical reasoning in gastrointestinal diagnosis. The results show that combining a reliable classifier with an LLM helps bridge the gap between what is visible in an endoscopic image and the type of explanation a clinician expects. When the classifier’s prediction is correct, most LLMs can generate structured and medically meaningful reasoning, including causes, symptoms, treatment options, lifestyle suggestions and follow-up care. However, the findings also make it clear that the entire pipeline depends heavily on upstream accuracy -  any classification error leads directly to flawed reasoning because LLMs cannot interpret images on their own and rely entirely on the predicted label.  

Our proposed method integrates DL with LLMs to bring gastrointestinal diagnosis closer to transparent, interpretable, and multi-step reasoning systems that resemble expert thinking. The pipeline illustrates how automated disease detection can be extended to clinically meaningful explanations, pointing toward decision-support tools that better align with real-world clinical needs. This work lays the foundation for developing reasoning models that are not only more informative but also more consistent with expert interpretive patterns. This creates a new direction for future experiments and contributes more effectively to clinical decision-making.

\section{Limitations and Future Work}
\label{limitations}
In our experiment some models, especially the larger ones produce more stable and clinically aligned explanations but none of them consistently achieve expert-level reasoning. Smaller models often miss contextual details, struggle with medical nuance, and show greater sensitivity to prompt design. Even the strongest models display variation across disease classes and prompt types, indicating that they still fall short of the consistency and reliability needed for real clinical use.

Future research should aim to make LLM-based reasoning more dependable and medically grounded. One direction is to strengthen the \textbf{DL\textsuperscript{3}M} pipeline by improving disease-prediction accuracy, incorporating uncertainty estimation, and enabling LLMs to handle ambiguous or borderline cases more effectively. Another important direction is the development of domain-adapted LLMs trained specifically on gastrointestinal data, allowing their reasoning to reflect expert-level knowledge rather than broad, general medical patterns. For models that behave inconsistently, structured fine-tuning and reinforcement learning with clinician-verified feedback may help improve stability and reliability. A more robust evaluation setup is also needed, one that goes beyond textual similarity to measure factual correctness, clinical validity, and risk awareness. Finally, future systems should work toward real-time clinical decision support, where the LLM can interact with clinicians, ask clarifying questions, and justify its recommendations in a transparent manner. These developments can help shift models currently exhibiting moderate or low clinical reasoning alignment into trustworthy tools for sensitive medical tasks such as stomach-infection assessment, polyp management, and broader GI disease reasoning.


\bibliography{cas-refs}

\end{document}